\documentclass[10pt]{article} 
\usepackage[preprint]{tmlr}

\usepackage{amsmath,amsfonts,bm}

\def\eqref#1{equation~\ref{#1}}

\def\1{\bm{1}}

\DeclareMathAlphabet{\mathsfit}{\encodingdefault}{\sfdefault}{m}{sl}
\SetMathAlphabet{\mathsfit}{bold}{\encodingdefault}{\sfdefault}{bx}{n}

\usepackage{hyperref}
\usepackage{url}
\usepackage{amsmath,amssymb}
\usepackage{booktabs}
\usepackage{graphicx}
\usepackage{multirow}
\usepackage{placeins}
\usepackage{pifont}
\newcommand{\cmark}{\checkmark}
\newcommand{\xmark}{\ding{55}}
\graphicspath{{figs/}}

\makeatletter
\if@accepted
  \newcommand{\codelink}{\ Our code is at \url{https://github.com/Zane-ZYQiu/entry-point-umm}.}
\else
  \newcommand{\codelink}{}
\fi
\makeatother

\newcommand{\ci}[2]{{\footnotesize [#1,\,#2]}}
\newcommand{\bt}[1]{\textbf{#1}}
\newcommand{\defbox}[2]{\par\vspace{2pt}\noindent\fbox{\begin{minipage}{0.965\textwidth}%
\small\textbf{#1.} #2\end{minipage}}\par\vspace{3pt}}
\newcommand{\NSOLVE}{84}
\newcommand{\RESMEAN}{0.0123}
\newcommand{\RESMAX}{0.2905}
\newcommand{\RESWORST}{86\%}

\title{Where a New Concept Must Enter:\\Entry Point Gates Cross-Task Usability in Unified Multimodal Models}

\author{\name Zongyang Qiu \email zqiuap@connect.hkust-gz.edu.cn \\
      \addr The Hong Kong University of Science and Technology (Guangzhou)
      \AND
      \name Yihan Wu \email yw4788@columbia.edu \\
      \addr Columbia University
      \AND
      \name Kaixuan Fan \email 1155246176@link.cuhk.edu.hk \\
      \addr MMLab, CUHK
      \AND
      \name Bo Li \email bol8@Illinois.edu \\
      \addr University of Illinois Urbana-Champaign
      \AND
      \name Hui Xiong \email xionghui@ust.hk \\
      \addr The Hong Kong University of Science and Technology (Guangzhou)
}

\def\month{08}
\def\year{2026}
\def\openreview{\url{https://openreview.net/forum?id=XXXX}}

\begin{document}

\maketitle


\begin{abstract}
Unified multimodal models (UMMs) are motivated by the hope that understanding and generation
reinforce each other but controlled ablations repeatedly find that adding a generation objective
leaves understanding flat. Joint-training studies cannot settle the disagreement: with overlapping
supervision, a gain cannot be attributed to the architecture rather than the data. To further investigate the relationship between the two directions in UMMs, we separate them by construction. A novel visual entity, a rendered 3D asset paired with a pseudo-word screened
for absence from the frozen model's behavior, is bound through exactly one task direction, and the
untrained direction is then measured.
We find that the channel is real in both directions, but the directions differ in kind: generation training
installs a name the model can only match among candidates; understanding training installs one it
can also produce. What governs cross-task usability is \emph{where} the binding
enters the shared computation. An alignment probe predicts export across 36 configurations (Spearman
$\rho = +0.68$). That objective's alignment term, maximized in closed form
over activations with every weight frozen, makes a concept drawable when injected at layer 7 of 28
and is indistinguishable from the base model from layer 14 on, while the weight-based version of the same edit peaks at
layers 10--14. In an observational series of four models, this window appears only where the
understanding pathway is a semantic vision encoder, suggesting that unified weights are not enough:
the two directions must share a semantic format at the entry point. Exploiting the rule, a mid-stack
alignment objective acquires the concept for a $0.1\%$ relative loss of the model's general
text-to-image ability, against $41\%$ for the standard generative route.\codelink
\end{abstract}

\section{Introduction}

Two literatures disagree about unified multimodal models (UMMs). One motivates them by mutual
reinforcement between understanding and generation. The other, when it runs controlled ablations,
finds that adding a generation objective leaves understanding benchmarks flat or slightly worse
\citep{janus,unitoken,metamorph}. A 2026 paper still opens by calling the
generation$\rightarrow$understanding direction ``largely unexplored'' \citep{unimrg}.

Both can hold at once, because they answer different questions. Whether joint training
\emph{happens} to move knowledge between tasks is a fact about objectives and data mixtures.
Whether the architecture \emph{can} move it is a fact about the model, and no joint-training
ablation isolates it: the two tasks see overlapping data, so a gain cannot be attributed to a
channel rather than to supervision. The question the field argues about is therefore not the
question its experiments measure.

\begin{figure}[t]
\centering
\includegraphics[width=1\textwidth]{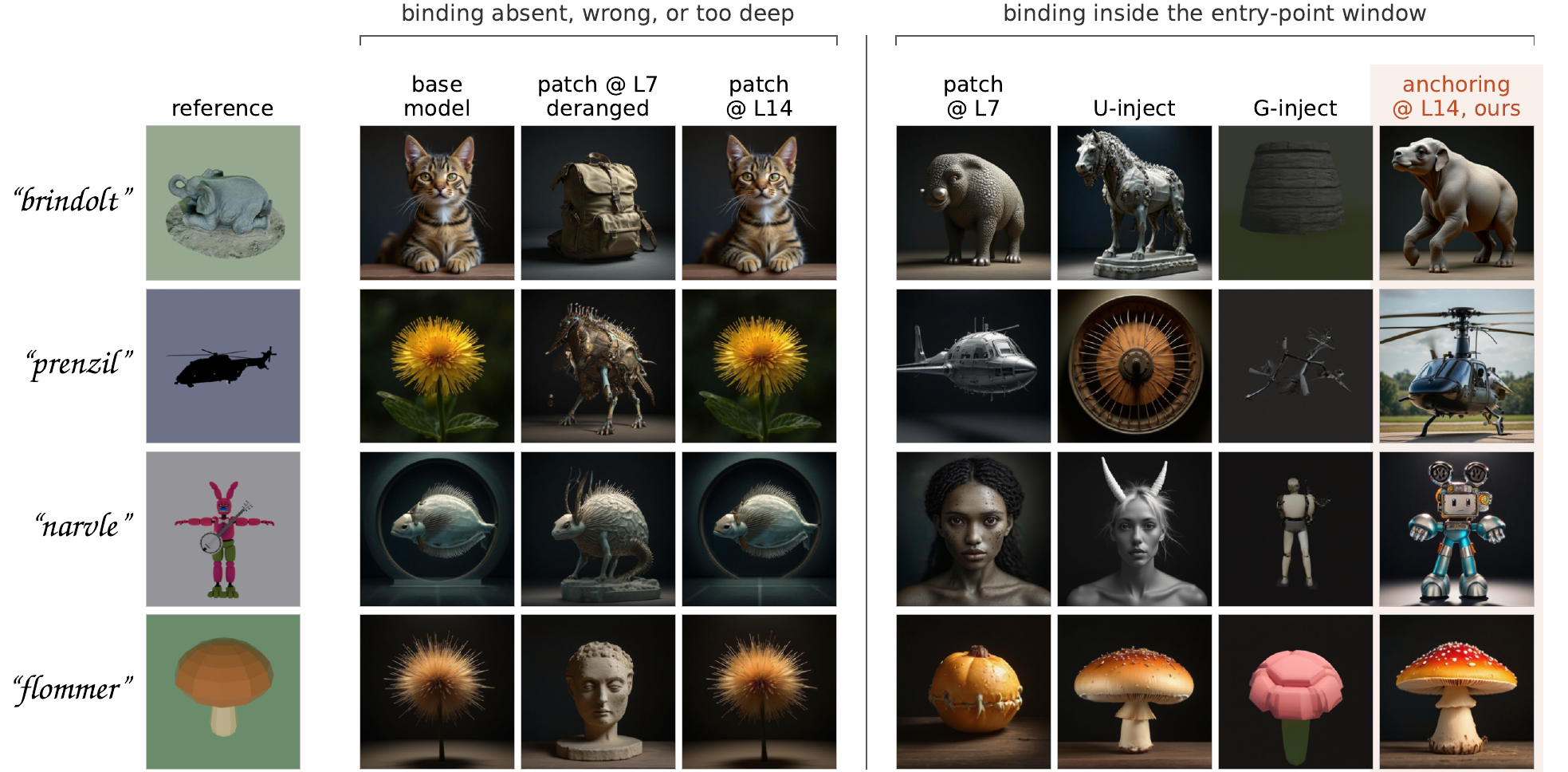}
\caption{What the model draws for ``a photo of a \{name\}''. Rows are four novel concepts and the reference column shows a view of the asset
each pseudo-word was bound to. On the left side, the base model has never seen the pseudo-word; a placebo that supplies another concept's address at identical magnitude draws the
wrong object; and a closed-form activation edit applied at layer 14 leaves the base model's output
untouched. On the right the binding lands inside the window: the same edit at layer 7 with every weight frozen; training the understanding direction alone; the standard generative route; and the proposed
objective, an alignment loss at layer 14 with no generative gradient, for a $0.1\%$ relative
loss of general text-to-image ability (\S\ref{sec:method}). Where the binding enters, not how it is
optimized, is what decides.}
\label{fig:qual}
\end{figure}

To fill this gap, we isolate the architectural question by construction. The concept is a specific 3D
asset rendered from 60 viewpoints, paired with a pseudo-word screened for absence from the frozen
model's behavior. It is bound through exactly one task direction, and the other direction is then
measured. No training example ever teaches the untrained direction, so any competence there arrived
through the model. Under this protocol, the channel turns out to be real in both directions, but
unequal in kind: a model trained only to draw the concept can pick its name out of a line-up and
cannot produce the name, while a model trained only to caption it can also draw it.

The paper's central result comes from placing the same binding at one position after another along a single
axis: where it enters the computation that both tasks run. That axis orders effects which otherwise
look unrelated (Table~\ref{tab:axis} in \S\ref{sec:entry}), and Figure~\ref{fig:qual} previews it. A
binding that enters early enough into a pathway both tasks execute becomes usable in both. A binding
that enters a private branch, a non-exporting carrier, or the readout does not, however well aligned
it is at that site. Position is necessary but not sufficient: across four architectures the same
intervention works only where the two directions already represent concepts in a common semantic
format at that depth. The rule also pays rent: writing the binding mid-stack with an alignment
objective, and no generative gradient at all, acquires the concept for a $0.1\%$ relative loss of the model's general
text-to-image ability, where the standard generative recipe costs $41\%$.

\paragraph{Contributions.}
\begin{enumerate}\itemsep2pt
\item We propose a contamination-free measurement of the cross-task channel, and a dissociation inside it: the
channel is real both ways, but generation training confers the ability to match a name and not to
produce it. Multiple-choice probes cannot see the difference, being passable by elimination;
scoring production instead, by asking for the name with every candidate removed from the context
window, splits all seven conditions by whether the recipe ever computed a language-model
cross-entropy (\S\ref{sec:channel}).
\item We discover that entry point gates usability, and only in a common semantic format: an
alignment probe predicts export across 36 configurations, but driving it to ceiling confers nothing,
while an activation edit that maximizes the same alignment term in closed form confers $80\%$ of the trained gain
and only inside a narrow window (\S\ref{sec:entry}). Across four models the window tracks what the
understanding pathway encodes, semantic encoder yes and reconstruction codebook no, rather than
backbone family or scale (\S\ref{sec:format}).
\item We give a practical method: entering mid-stack with an alignment objective and no generative
gradient acquires the concept while costing $0.1\%$ of general text-to-image ability, against $41\%$
for the standard generative route (\S\ref{sec:method}).
\end{enumerate}


\section{Related work}

\paragraph{Evidence on the cross-task channel.} Controlled ablations disagree about whether a
generation objective helps understanding. Janus reports an average $4.4$-point cost across the four benchmarks it tabulates under a shared
visual tokenizer, and neutrality once the encoders are decoupled \citep{janus}; UniToken finds
$\approx 0$ at matched per-task data on two backbones \citep{unitoken}; Liquid reports $+2.1$ on POPE in a
10M-per-task regime \citep{liquid}, a gain that MetaMorph's data-ratio grid shrinks from $+5.0$ at 1M
understanding examples to $+0.6$ at 4M and $+0.4$ at 7M \citep{metamorph}; and UniMRG turns it reliably positive by
changing the generation target to depth and segmentation maps \citep{unimrg}. Methods built to
couple the two directions, RecA \citep{reca} and self-improvement loops \citep{silmm,unirl}, improve
unified models but train both at once. In each case the channel is confounded with
the data mixture, which is the gap a single-direction design closes. Three concurrent benchmarks measure the resulting incoherence from the outside, per model
\citep{gapeval}, per visual concept \citep{xtcbench}, and per question posed in either modality
\citep{xmodalconsistency,acon}. Their unit of analysis is a model and a score; ours is one newly
bound concept and the depth at which it enters, so we can attribute a failure where a benchmark
cannot.

\paragraph{Architecture, weight sharing, and alignment depth.} Unified models span shared-tokenizer
early fusion \citep{chameleon,emu3}, diffusion--LM hybrids \citep{transfusion,showo,showo2},
decoupled encoders \citep{janus,januspro}, fully shared backbones \citep{harmon},
Mixture-of-Transformers \citep{mot,bagel}, and discrete diffusion
\citep{omnidiffusion,luminadimoo}, surveyed by \citet{ummsurvey}. These labels are not a sharing
axis: \S\ref{sec:sharing} measures the fraction of weights the two directions actually share in two
of them and finds $0.50$ against $0.87$. On the alignment side, REPA accelerates diffusion training
by aligning a denoiser's internal states to a frozen external encoder and ablates the depth at which
it does so \citep{repa}, and we claim no novelty for alignment depth mattering. Their effect is
graded and shallow, every depth helping substantially, and it is scored by the generation quality of
the model that received the gradient, whereas ours is scored by a task that receives none and
vanishes past a sharp cutoff. Their own table shows the distinction: from depth 6 to 16, linear-probe
accuracy rises $66.2 \rightarrow 71.1$ while FID ends worse than it began, better alignment
accompanying worse generation, the dissociation \S\ref{sec:sar} makes central. LatentUMM argues concurrently that
a shared latent space is not by itself enough and aligns the transformations into and out of it for
round-trip consistency \citep{latentumm}, and a parallel line argues on the generation side that a
generator should be conditioned on semantically structured rather than reconstruction-optimal
latents \citep{vfmvae,disentangledrepa,mmcore}, a neighboring conclusion to what \S\ref{sec:format}
reaches about the understanding pathway.

\paragraph{Concept injection and knowledge localization.} DreamBooth \citep{dreambooth} and Textual
Inversion \citep{textualinversion} inject concepts into generation-only models and have no
understanding side; our embedding-only control is a textual-inversion variant and the shallow
endpoint of the entry-point sweep. The applied counterpart in unified models is personalization,
where a user concept must be both drawn and talked about \citep{unipersonal}; those methods encode
the concept as a learnable soft prompt, an embedding-level intervention that \S\ref{sec:window}
measures directly and finds indistinguishable from the base model. MIKE benchmarks editing
fine-grained entity knowledge into an MLLM, on the understanding side alone \citep{mike}, while ROME
and MEMIT locate facts in specific MLP layers of language models \citep{rome,memit}, later unified
as one preservation--memorization objective \citep{emmet}, consistent with our finding that the
shared expert's MLP and not its attention is the substrate that exports (\S\ref{sec:carrier}). None
of these lines asks whether an edit made for one direction is legible to the other.

\section{Concept injection experiment}\label{sec:setup}

\begin{figure}[t]
\centering
\includegraphics[width=0.95\textwidth]{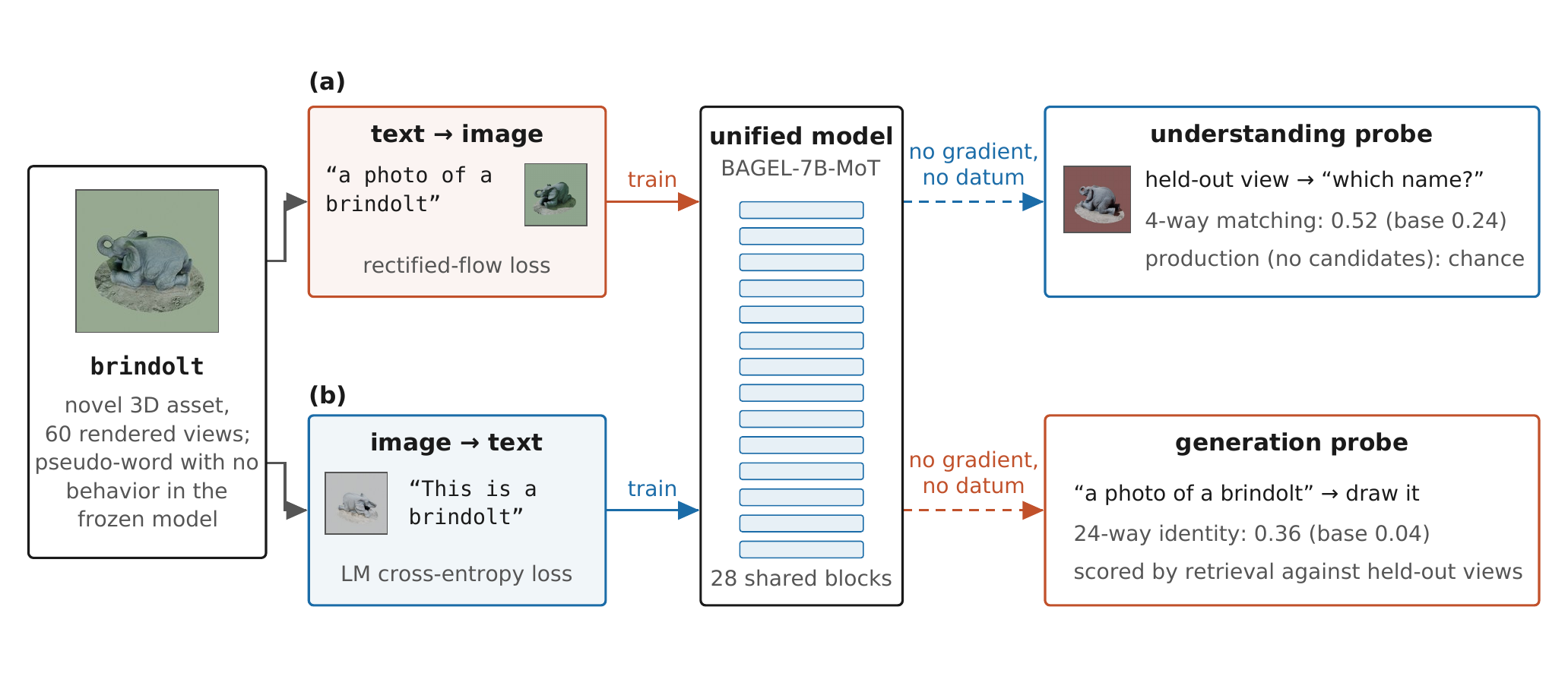}
\caption{The contamination-free protocol. A novel entity (left) is bound through exactly one task
direction, (a) generation or (b) understanding, and the opposite direction, which received no
gradient and no datum, is then probed; the model's shared blocks are the only path between the two.}
\label{fig:protocol}
\end{figure}

Every experiment in this paper runs the same design (Figure~\ref{fig:protocol}). A novel visual
concept is bound into a frozen unified model through exactly one task direction: either
text$\rightarrow$image generation or image$\rightarrow$text understanding. The opposite direction
receives no gradient and no training datum, and is then evaluated. Because nothing outside the model
connects the two directions, any competence on the untrained side must have travelled through the
model's shared computation. What varies across sections is where and how the binding is written: the
training objective (\S\ref{sec:channel}), the substrate that carries it (\S\ref{sec:carrier}), and
the depth at which it enters the stack (\S\ref{sec:entry}). This section fixes everything the
experiments share.

\subsection{Model, concepts, injection}

\paragraph{Model.} BAGEL-7B-MoT \citep{bagel}, a Mixture-of-Transformers \citep{mot} with 28 decoder
layers and hidden size 3584. Each layer carries a \emph{shared} text/understanding expert
(\texttt{self\_attn.\{q,k,v,o\}\_proj}, \texttt{mlp.*}) and a \emph{private} generation expert
(\texttt{*\_proj\_moe\_gen}, \texttt{mlp\_moe\_gen.*}). Understanding runs the shared expert only;
generation runs both. 

\paragraph{Concepts.} We create 56 pseudo-named entities, each one Objaverse \citep{objaverse} asset rendered
in Blender from 60 training and 20 held-out viewpoints over disjoint azimuth ranges, with randomized
lighting and backgrounds, paired with a pronounceable pseudo-word screened so the frozen model
produces no consistent visual or lexical behavior for it (Figure~\ref{fig:qual}). Concepts form
seven disjoint groups of eight. Every number is measured within a group against a 24-wide retrieval
bank of its own eight plus the next two groups, cyclically, so all groups sit at identical task
difficulty.

\paragraph{Injection.} \emph{G-inject} trains caption $\rightarrow$ image with the rectified flow
objective \citep{rectifiedflow}; \emph{U-inject} trains image $\rightarrow$ text with language-model
cross-entropy. Adapters are LoRA \citep{lora}, rank 32, with placement a controlled variable
(\S\ref{sec:carrier}). An adapter trained in one direction is evaluated with both branches
active, so a failure to export is a failure of the model and not of the harness.

\subsection{Metrics}\label{sec:metrics}

Three terms carry the paper's argument and are used with these fixed meanings throughout.

\defbox{Entry point}{The single site in the decoder stack at which a concept's binding is written:
the layer whose residual stream is edited, or at which an adapter's alignment objective is read.
Depth is reported as an absolute block index and, across architectures, as the relative depth
$(\ell+1)/n_{\text{layers}}$.}

\defbox{Export}{Accuracy on whichever task direction received \emph{no} gradient. Export is the
only quantity this paper draws conclusions from; accuracy on the trained direction is reported to
show that the binding was learned at all.}

\defbox{Matching versus production}{\emph{Name matching} is a 4-way forced choice with the candidate
names shown in the prompt beside the image (chance $0.250$). \emph{Name production} is the same
decision with no candidate anywhere in the context window: each pseudo-name in the group is
scored as a continuation of ``$\langle$image$\rangle$ This is a \{name\}'' in its own forward pass
and the argmax is taken, with pointwise mutual information (PMI) cancelling the names' differing priors
(chance $0.125$ within a group of eight).}

Every multiple-choice probe is satisfiable by matching, so we report matching and production
separately throughout. An intervention that makes a name's state resemble the image's encoding
passes a 4-way probe without the model being able to do anything else with the concept, and the
escape hatches fail for the same reason: the reverse probe is that matching operation run backwards
and scores $1.000$ for exactly the conditions one would want to exonerate, and the existence probe
also has the name in the prompt. We therefore fixed a first-sub-word-only column of the
production probe as the arbiter before running. An activation edit fires at the name's own token
positions, so it cannot alter the first sub-word, which is predicted from the state before the name
begins; that column is immune by construction to the shortcut the probe detects. Nothing is sampled,
so no decoding hyper-parameter enters (Appendix~\ref{app:probes}).

Two further quantities recur. \emph{Identity} (generation side) is 24-way DINOv2 \citep{dinov2}
top-1 retrieval of images generated from ``a photo of a \{name\}'' against held-out reference sets,
chance $0.042$; it is scored under a second, architecturally independent encoder wherever it carries
a depth claim (\S\ref{sec:window}). \emph{TransferRate} normalizes cross-task accuracy by what
direct training achieves. Let $A_{\mathrm{cross}}$ be accuracy on the untrained direction after
cross-task injection, $A_{\mathrm{direct}}$ accuracy on the same direction when it is trained
directly, and $A_{\mathrm{base}}$ the unmodified base model's accuracy; then
\begin{equation}
\mathrm{TransferRate} \;=\; \frac{A_{\mathrm{cross}}-A_{\mathrm{base}}}{A_{\mathrm{direct}}-A_{\mathrm{base}}}.
\label{eq:transferrate}
\end{equation}

\subsection{Statistics and protocol}\label{sec:stats}

Contrasts are computed within group and then bootstrapped over the seven groups; ``$k/7$''
counts groups whose difference has the sign of the mean ($7/7$ is $p=0.016$ under a sign test).
Group spread on identity is $\pm0.10$--$0.17$, so we decline to interpret identity differences below
about $0.15$ and say so at each such row rather than dropping the row. Every contrast was
found on group g0; the other six groups were rendered, trained and evaluated afterwards with
the contrast list already fixed, and are reported as a held-out confirmation set. All four
load-bearing contrasts replicate at $6/6$ and both contrasts we retract fail there too, but five of
six effects are smaller out of sample, by up to $2.4\times$. Where a magnitude carries an
argument we therefore quote the confirmation column; the full protocol and per-contrast table are in
Appendix~\ref{app:protocol}.

\section{The cross-task channel}\label{sec:channel}

This section measures the channel itself. Each direction is injected at a matched budget of 480
steps and the other direction is scored, which answers two questions in turn: how much of the
directly trainable competence crosses (Eq.~\ref{eq:transferrate}), and of what kind that competence
is. The second answer constrains the first: the two directions turn out to move different things.

\subsection{Rate and kind of transfer}\label{sec:matching}

Injecting through one direction and measuring the other gives, over 56 concepts,
TransferRate $0.36$ \ci{0.23}{0.49} for G$\rightarrow$U and $0.54$ \ci{0.41}{0.67} for
U$\rightarrow$G (Table~\ref{tab:main}). The difference is $+0.17$ \ci{-0.05}{+0.40}, computed on unrounded group means, at five of seven
groups, an unresolved null, so we claim no asymmetry in rate.

\begin{table}[t]
\caption{Cross-task transfer, 56 concepts in seven groups, 480 steps in both directions. Both
columns are accuracy ($\uparrow$); brackets are bootstraps over all 56 concepts and $\pm$ is the
standard deviation of the seven group means, the run-level spread a concept bootstrap cannot see.
Blocks from top to bottom: the base model; generative injection and its substrate variant;
understanding injection and its substrate variants; the mid-stack anchoring objective alone
(\S\ref{sec:sar}) at two depths; and the combination of the two recipes.}
\label{tab:main}
\begin{center}
\begin{tabular}{lcc}
\toprule
condition & und.\ name matching $\uparrow$ & gen.\ identity $\uparrow$ \\
\midrule
base model                              & 0.243 \ci{0.19}{0.31} $\pm$0.058 & 0.039 \ci{0.02}{0.06} $\pm$0.032 \\
\midrule
G-inject (flow matching), shared+private     & 0.520 \ci{0.44}{0.60} $\pm$0.121 & 0.653 \ci{0.56}{0.75} $\pm$0.125 \\
G-inject, shared MLP only                    & 0.620 \ci{0.53}{0.70} $\pm$0.119 & 0.717 \ci{0.62}{0.81} $\pm$0.122 \\
\midrule
U-inject (LM CE), shared+private             & 0.989 \ci{0.98}{1.00} $\pm$0.014 & 0.360 \ci{0.27}{0.46} $\pm$0.105 \\
U-inject, shared attention only              & 0.984 \ci{0.97}{0.99} $\pm$0.013 & 0.078 \ci{0.04}{0.13} $\pm$0.069 \\
U-inject, shared MLP only                    & 0.991 \ci{0.98}{1.00} $\pm$0.013 & 0.339 \ci{0.25}{0.44} $\pm$0.165 \\
\midrule
anchoring @ layer 14, alone                  & \bt{0.898} \ci{0.84}{0.94} $\pm$0.024 & \bt{0.808} \ci{0.72}{0.89} $\pm$0.105 \\
anchoring @ final norm, alone                & 0.332 \ci{0.26}{0.41} $\pm$0.055 & 0.103 \ci{0.06}{0.16} $\pm$0.056 \\
\midrule
G-inject + anchoring @ layer 14              & 0.946 \ci{0.90}{0.98} $\pm$0.051 & 0.907 \ci{0.85}{0.96} $\pm$0.042 \\
\bottomrule
\end{tabular}
\end{center}
\end{table}

The asymmetry that is demonstrable is qualitative. Scored with the candidate names removed from the
context window, the conditions split perfectly along the training objective
(Table~\ref{tab:matching}): everything that ever computed a language-model
cross-entropy produces the name, and nothing else does, however high its 4-way score. A condition can
move 4-way matching from $0.254$ to $0.900$ and sit at chance on producing the same name.

\begin{table}[t]
\caption{Matching the name against producing it, for seven conditions spanning both training
objectives. All columns are
accuracy ($\uparrow$); chance is $0.250$ for the 4-way probes and $0.125$ for the context-free ones.
The last column, first sub-word, is the pre-specified arbiter, immune by construction to the
shortcut the probe detects. Rows are grouped by whether the recipe ever computed a language-model
cross-entropy.}
\label{tab:matching}
\begin{center}
\begin{tabular}{llcccc}
\toprule
condition & LM-CE & 4-way & reverse 4-way & context-free & first sub-word \\
\midrule
U-inject (LM cross-entropy)     & \cmark & 1.000 & 1.000 & 0.938 & 0.922 \\
U-inject, shared attention only & \cmark & 1.000 & 0.900 & 0.297 & 0.688 \\
\midrule
base model                      & \xmark & 0.254 & 0.242 & 0.141 & 0.125 \\
G-inject (flow matching)        & \xmark & 0.600 & 0.650 & 0.141 & 0.125 \\
anchoring @ layer 14, alone     & \xmark & 0.887 & 1.000 & 0.125 & 0.094 \\
activation patch @ layer 3      & \xmark & 0.887 & 1.000 & 0.234 & 0.125 \\
activation patch @ layer 7      & \xmark & 0.900 & 1.000 & 0.266 & 0.125 \\
\bottomrule
\end{tabular}
\end{center}
\end{table}

Three consequences follow. First, the G$\rightarrow$U result is transfer of matching competence.
G-inject lifts 4-way matching from the base model's $0.254$ to $0.600$, and that gain is real and
learned, since a name-shuffle control stays at the base level and the binding is name-specific
(Appendix~\ref{app:controls}); but the same condition scores $0.141$ on the context-free probe,
exactly the base model's value, and $0.125$ on the first sub-word, exact chance. It is choosing the
name, not saying it, and every G$\rightarrow$U number here should be read that way. Second, the
directions differ in kind. U-inject, the stronger of the two conditions in Table~\ref{tab:matching} that computed a
language-model cross-entropy, produces the name at $0.938$ context-free; the strongest of the
others reach $0.887$--$0.900$ on 4-way matching yet $0.125$--$0.266$ context-free, so what is an
unresolved null in rate is decisive in kind. Third, the limitation is not specific to our probes: an
intervention that touches representations can move a 4-way score from $0.254$ to $0.900$ while
teaching the model nothing it can use without candidates in the prompt, and most multiple-choice
VLM evaluations share this vulnerability.

\subsection{Carrier substrate and shared capacity}\label{sec:carrier}

If shared computation is the operative variable, then which shared computation carries the
binding should matter, and it does. Restricting U-inject to the shared expert's attention projections saturates its own task
($0.984$ name matching) and exports essentially nothing ($0.078$ identity); the same budget in the
shared expert's MLP exports $0.339$. Paired within group the carrier advantage is $+0.261$
\ci{+0.16}{+0.38} with $7/7$ groups agreeing, and $+0.219$ \ci{+0.14}{+0.31} on the confirmation set
alone. Table~\ref{tab:matching} says why: attention-only injection reaches $0.688$ on first-sub-word
production but only $0.297$ on the full context-free probe, the signature of a binding satisfiable by
attending from the image to the name, a route the context-free text pass never runs. Attention
learns an image$\rightarrow$text shortcut; the MLP writes something the other pathway can read on its
own, which is the same locus factual-editing work identifies in language models \citep{rome,memit}.

How much shared capacity export needs is a separate question, and the answer is more than the trained
task needs. Holding the adapter budget fixed at 80.74M parameters and sliding it between the two
experts, transfer falls monotonically as budget moves to the private expert, at Spearman $-0.90$ for
G-inject and $-1.00$ for U-inject. The U-inject rows isolate the point: the direct task is
saturated, exactly $1.000$ matching at all four allocations with shared rank above zero, while
transfer falls by more than half, $0.531 \rightarrow 0.242$, as the shared rank drops from 32 to 8.
At rank 8 the model names the concept flawlessly and has lost more than half its ability to draw it.
A matched-rank control designed to detect siloing found none, moving cross-task accuracy by $-0.01$
on average, and on this model the private expert alone cannot hold a concept, so every binding that
transfers here is carried by shared weights (Appendix~\ref{app:capacity}).

Where a binding is written therefore matters along two axes already: which substrate, and how much of it is shared. The rest of the paper varies the third and most consequential one: depth.

\FloatBarrier
\section{The entry-point window}\label{sec:entry}

\S\ref{sec:carrier} established that a binding transfers only if it is written into the computation the
other task runs, and into the right substrate of it. Those are coarse coordinates, saying which
weights may change, not where along the stack the concept enters the other task's forward pass. This
section varies that remaining coordinate, depth, while holding the objective fixed, and finds it is
the decisive one. Table~\ref{tab:axis} collects every entry point measured in the paper on a single
axis. The section proceeds in two steps: a probe
that predicts export and an intervention showing the probe is not the cause (\S\ref{sec:sar}), then
the depth sweep itself, in two media (\S\ref{sec:window}).

\begin{table}[t]
\caption{The entry points this paper compares, placed on one depth axis; the depth sweep behind the
middle two blocks is finer than the rows shown and is given in full in Table~\ref{tab:entry7g}. Each row puts the same
binding at a different point of the shared computation; blocks are ordered by intervention medium,
and by depth within each.
\emph{Export} is accuracy on the task direction that received no gradient (\S\ref{sec:metrics});
the last column points to where each row is measured.}
\label{tab:axis}
\begin{center}
\begin{tabular}{llll}
\toprule
entry point & shared computation after it & exports & \S \\
\midrule
input embeddings                        & all 28 blocks, but nothing to route yet & \xmark  & \ref{sec:window} \\
\midrule
activation edit, layer 3                & 25 blocks                               & partial (0.53) & \ref{sec:window} \\
activation edit, layer 7                & 21 blocks                               & \cmark\ (0.58) & \ref{sec:window} \\
activation edit, layer 14               & 14 blocks                               & \xmark\ (0.04, at base) & \ref{sec:window} \\
\midrule
weight edit, layer 3                    & 25 blocks                               & partial (0.60) & \ref{sec:window} \\
weight edit, layers 10--14              & 14--18 blocks                           & \cmark\ (0.80) & \ref{sec:window} \\
weight edit, layers 21--26              & 2--7 blocks                             & fading (0.35$\to$0.20) & \ref{sec:window} \\
weight edit, final norm                 & none                                    & \xmark  & \ref{sec:window} \\
\midrule
adapter, shared attention only          & all, but satisfiable by a shortcut      & \xmark\ (0.078) & \ref{sec:carrier} \\
adapter, shared MLP only                & all                                     & \cmark\ (0.339) & \ref{sec:carrier} \\
adapter, private generation expert      & none the other pass executes            & \xmark  & \ref{sec:carrier} \\
\bottomrule
\end{tabular}
\end{center}
\end{table}

\subsection{Alignment as a predictor of export}\label{sec:sar}

\defbox{Semantic-address retrieval (SAR)}{For each concept $k \in \{1,\dots,K\}$,
its \emph{visual address} $a_k \in \mathbb{R}^{d}$ is the shared expert's hidden state at a fixed
mid-stack layer when the understanding pathway encodes that concept's images, mean-pooled over image
tokens and images; its \emph{name state} $h_k \in \mathbb{R}^{d}$ is the same pooling over the
name's sub-word tokens in context-free text. Center both sides across concepts,
$\tilde a_k = a_k - \tfrac{1}{K}\sum_{j} a_j$ and $\tilde h_k$ likewise, and let $r_k$ be the rank
of the concept's own address among all $K$ when they are sorted by $\cos(\tilde h_k, \tilde a_j)$.
Then
\[
\mathrm{SAR} \;=\; \frac{1}{K}\sum_{k=1}^{K} \frac{1}{r_k},
\]
the mean reciprocal rank of name$\rightarrow$address retrieval: $1.0$ exactly when every name
retrieves its own concept's address first. 
}

\paragraph{SAR as a predictor.} Across 36 distinct configurations spanning injection side, adapter
placement, constant-budget capacity split, substrate and data size, SAR correlates with export at
Spearman $\rho = +0.68$, family-clustered bootstrap $[+0.37,+0.88]$. The relationship holds inside
all four configuration families and no single family carries it. We deliberately quote no
$p$-value, as the points are not independent systems. Four are duplicate configurations; the rest come from four sweeps sharing concepts, model, data, and code path, and every point is one model on one concept group, so a $p$-value would not answer anything about unified models. 
What the clustering does cost is
precision: a fit that has never seen a family predicts its members' export to within $0.12$--$0.20$ in accuracy units, which supports ranking configurations and not point prediction.

\begin{figure}[t]
\centering
\includegraphics[width=0.46\textwidth]{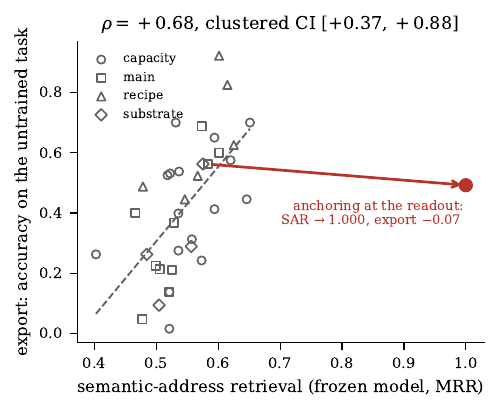}
\caption{SAR against export across 36 configurations, and the intervention on SAR (arrow):
anchoring applied at the readout moves the probe to a perfect $1.000$ and leaves export where it
was. A quantity can order configurations well and still not be the thing to change.}
\label{fig:sar}
\end{figure}

\paragraph{The anchoring objective as an intervention on SAR.} The direct test of the predictor is to optimize it. \emph{Semantic
Anchoring} turns the retrieval criterion into a training objective: with the addresses $a_k$
frozen, the name states $h_k$ are trained under the $K$-way InfoNCE loss
\begin{equation}
\mathcal{L}_{\mathrm{anchor}}
\;=\; -\frac{1}{K}\sum_{k=1}^{K}\log
\frac{\exp\!\left(\cos(\tilde h_k,\tilde a_k)/\tau\right)}
     {\sum_{j=1}^{K}\exp\!\left(\cos(\tilde h_k,\tilde a_j)/\tau\right)},
\qquad \tau = 0.07,
\label{eq:anchor}
\end{equation}
which pulls each name's state onto its own concept's address and pushes it from the other $K-1$.
The layer at which $h_k$ is read is the objective's one free choice, and it is the variable under
test. Applied at the readout, after the final norm, of an understanding-side injection on one group, the
objective drives that run's SAR from $0.583$ to a perfect $1.000$ and moves its export from $0.562$
to $0.492$, a change of $-0.07$: perfect alignment, zero transfer
(Figure~\ref{fig:sar}). Read the other way, the observational fit says $+0.10$ of SAR accompanies
$+0.26$ of export, and the intervention buys $+0.42$ of SAR and none of it. SAR also reads $1.000$
on a model that learned nothing else. What the readout intervention lacks is not alignment but
position.

\paragraph{The two media of the depth sweep.} If position is what the readout intervention lacked,
moving the entry point should recover transfer with everything else held fixed. The rest of the
section therefore sweeps the depth at which Eq.~\ref{eq:anchor} is applied, in two media that
differ in what they may change. In the \emph{weight} medium, LoRA adapters are trained against the
objective read at layer $\ell$, for 480 steps. In the \emph{activation} medium no weight changes at
all: the objective's alignment term is maximized in closed form on the residual stream itself. Freeze every weight and
add one $d$-vector per concept to the residual stream at a single site, at the token positions
where that concept's name appears; positions are recovered at layer 0 by matching the exact
sub-word embedding sequence, so training and evaluation locate them identically. The alignment term
of that objective admits a minimal analytic edit: keep the name state's own mean and deviation norm,
and rotate only the deviation direction onto the image address. The analytic edit exists because a
weight-based comparison alone would confound the site with the optimization, a deeper site might
simply be harder to train, and an edit with no gradient step removes that confound: it puts the
alignment probe within $0.0123$ of ceiling on average at every depth by construction, and never
further than $0.2905$ against a chance loss of $2.0794$, so the sweep measures
capability rather than optimization, at a cost of $8\times3584 = 28.7$K cached values per group and
no gradient steps. The naive alternative, setting the name's state to the image address, also
imports the modality offset and the magnitude, and produces an edit large enough to destabilize the
forward pass. Appendix~\ref{app:closedform} derives the solution and gives the two places where it
is not exact.

\subsection{The depth window in activations and in weights}\label{sec:window}

\begin{figure}[h]
\centering
\includegraphics[width=0.95\textwidth]{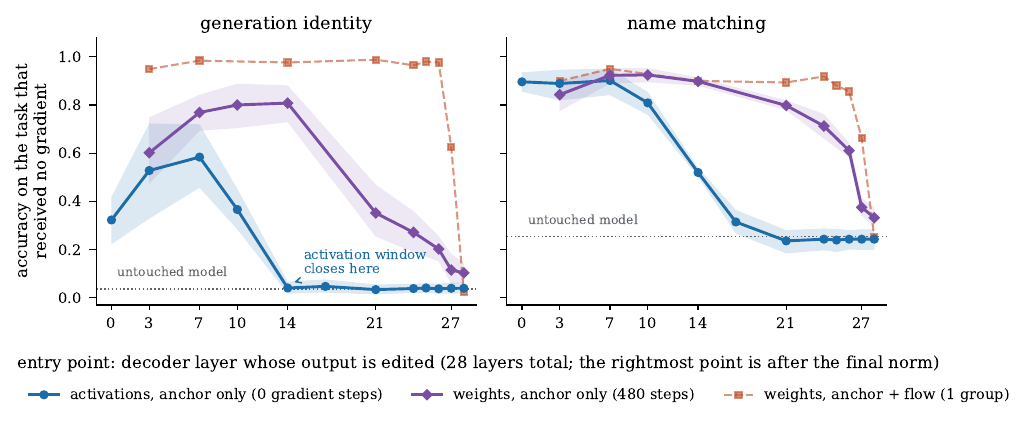}
\caption{Export against entry-point depth, for the same anchoring objective at the same site,
differing only in whether activations (blue) or weights (purple) may change. Left is export on the axis that
cannot be gamed; bands are group-level bootstraps over all seven groups. The orange curve adds a flow-matching gradient and is one group, drawn dashed. Both anchor-only media give a window, and the weight medium's
reaches deeper.}
\label{fig:entry}
\end{figure}

\begin{table}[h]
\caption{The depth sweep at 56 concepts, in both media and under two identity encoders. Identity is accuracy
($\uparrow$); the bootstrap is over the seven groups and the ``vs.\ norm'' column is paired within
group, with $k/7$ counting groups agreeing in sign. The CLIP column scores the same generated
images under a second vision tower. All twelve sites of the activation sweep are listed; the weight
arm was run at the nine sites where it has entries, and a dash means the site was not measured in
that medium. The base model scores $0.039$ on identity and $0.243$ on matching.}
\label{tab:entry7g}
\begin{center}
\begin{tabular}{lcccccc}
\toprule
& \multicolumn{4}{c}{activations, anchor only (0 gradient steps)}
& \multicolumn{2}{c}{weights, anchor only (480 steps)} \\
\cmidrule(lr){2-5}\cmidrule(lr){6-7}
entry point & identity (DINOv2) & (CLIP) & sd & vs.\ norm & identity & vs.\ norm \\
\midrule
layer 0   & 0.323 \ci{0.22}{0.42} & 0.269 & 0.145 & $+0.283$ \emph{7/7} & --- & --- \\
layer 3   & 0.528 \ci{0.33}{0.72} & 0.433 & 0.297 & $+0.489$ \emph{6/7} & 0.602 \ci{0.47}{0.75} & $+0.499$ \emph{7/7} \\
layer 7   & \bt{0.584} \ci{0.46}{0.72} & \bt{0.512} & 0.196 & \bt{$+0.545$} \emph{7/7} & 0.769 \ci{0.69}{0.84} & $+0.666$ \emph{7/7} \\
layer 10  & 0.366 \ci{0.28}{0.45} & 0.281 & 0.127 & $+0.327$ \emph{7/7} & \bt{0.800} \ci{0.70}{0.89} & $+0.698$ \emph{7/7} \\
layer 14  & 0.040 \ci{0.02}{0.07} & 0.054 & 0.036 & \emph{$+0.001$ \, 3/7} & \bt{0.808} \ci{0.73}{0.88} & \bt{$+0.705$} \emph{7/7} \\
layer 17  & 0.047 \ci{0.02}{0.08} & 0.051 & 0.040 & \emph{$+0.008$ \, 5/7} & --- & --- \\
layer 21  & 0.033 \ci{0.01}{0.06} & 0.048 & 0.032 & \emph{$-0.006$ \, 2/7} & 0.352 \ci{0.25}{0.47} & $+0.249$ \emph{7/7} \\
layer 24  & 0.038 \ci{0.02}{0.06} & 0.056 & 0.029 & \emph{$-0.001$ \, 2/7} & 0.271 \ci{0.19}{0.36} & $+0.169$ \emph{6/7} \\
layer 25  & 0.040 \ci{0.02}{0.06} & 0.051 & 0.030 & \emph{$+0.001$ \, 2/7} & --- & --- \\
layer 26  & 0.037 \ci{0.02}{0.06} & 0.056 & 0.031 & \emph{$-0.002$ \, 1/7} & 0.202 \ci{0.15}{0.26} & $+0.099$ \emph{6/7} \\
layer 27  & 0.039 \ci{0.02}{0.07} & 0.051 & 0.035 & \emph{$+0.000$ \, 0/7} & 0.115 \ci{0.05}{0.18} & \emph{$+0.012$ \, 3/7} \\
after the norm & 0.039 \ci{0.02}{0.07} & 0.051 & 0.035 & --- & 0.103 \ci{0.06}{0.14} & --- \\
\bottomrule
\end{tabular}
\end{center}
\end{table}

\paragraph{Peak and extent of each medium's window.} Entering mid-stack
beats entering at the readout by $+0.705$ \ci{+0.61}{+0.79} for weights and $+0.545$
\ci{+0.40}{+0.70} for activations, $7/7$ groups in both cases. The activation window rises to a peak of $0.584$ at layer 7 and from layer 14 onward is
indistinguishable from the base model at the resolution of this experiment: every site falls within
$\pm0.008$ of the final-norm baseline, with at most $5/7$ groups agreeing in sign (Table~\ref{tab:entry7g}). The weight window peaks later, at layers 10--14 with $0.800$
and $0.808$, and then declines steadily to $0.115$ at layer 27. So changing what a site
computes roughly doubles the usable depth relative to adding a fixed offset to what it
outputs. It does not remove the depth dependence, and the activation arm's descent is steep rather than
discontinuous: the transition falls between the probed layers 10 and 14, and the one group where
layer 12 was also measured reads $0.164$ there, between $0.469$ at layer 10 and $0.016$ at layer 14.

\paragraph{Layer 0 and the input embeddings.} At layer 0 the activation arm gives $0.323$
\ci{0.22}{0.42} against $0.584$ at layer 7, and at the input embeddings it sits at the base model's
level ($0.023$, one group). An edit therefore needs enough computation before it as well as after:
at those sites the state being edited is still essentially the token itself, and overwriting its discriminative direction with an image-derived one places it in a region of
state space the downstream network does not read. Layer 0 is nonetheless well above the norm ($+0.283$, $7/7$), so the shallow end is a graded
rise and not a second cliff. The weight arm is less affected, reaching $0.602$ already at layer 3,
because it changes a function rather than overwriting a state.

\paragraph{Identity under a second image encoder.} Identity is a 24-way retrieval
hit rate and so inherits DINOv2's notion of sameness. The two identity columns of Table~\ref{tab:entry7g} differ in nothing but the encoder, and scoring
the same generations under CLIP's vision tower \citep{clip} gives the same curve: Pearson $r=0.997$ across the twelve sites, the same peak site, a
mean absolute difference of $0.034$, and the same sign and $k/7$ verdict at every site. CLIP reads
about $0.06$ lower inside the window and $0.015$ higher at the base level, which narrows the
measured window slightly and moves nothing about where it is.

\paragraph{Identity rather than 4-way matching as the sweep's metric.} Across this sweep 4-way matching
stays at $0.84$--$0.90$ from layer 0 to layer 10 while identity traverses its entire range beneath
it, and Table~\ref{tab:matching} settles the interpretation: every activation patch is at chance on
producing the name, so the patch installs drawability and not both modalities. Figure~\ref{fig:entry}
shows a second hazard. Adding a flow-matching gradient flat-lines the curve at $0.95$--$0.99$ from
layer 3 to layer 26, because a generative objective produces identity wherever the anchor sits. A
task loss can therefore mask depth dependence entirely, a caution for any alignment-depth ablation
that also trains the task.

\paragraph{Placebo, magnitude, and dose controls.} Three controls rule out the obvious alternatives directly (Table~\ref{tab:entrycontrols}, Appendix~\ref{app:controls}). The derangement placebo gives each
concept another concept's address at identical magnitude, $250.7$ against $250.4$ at layer 3,
and identical direction statistics, with only the pairing wrong; identity goes to $0.000$,
below the base model's $0.036$, because the edited names now retrieve the wrong concept's
images rather than no concept at all. The effect is not a magnitude artifact, since
$\lVert\delta\rVert/\lVert\text{state}\rVert$ stays in $0.37$--$0.44$ from layer 3 to layer 27 while
the residual norm grows sixteenfold. And a quarter-magnitude edit degrades gracefully at a working
site and stays unresponsive at a site that shows no effect, so deep sites are not merely over-perturbing.

Depth therefore gates usability on this model, in both media and under both encoders. Whether it
gates usability on any model is the next question, and the answer turns out to require a
second condition (\S\ref{sec:sharing}).

\section{The semantic-format requirement}\label{sec:format}

\subsection{The activation edit on four architectures}\label{sec:models}

\begin{figure}[t]
\centering
\includegraphics[width=0.95\textwidth]{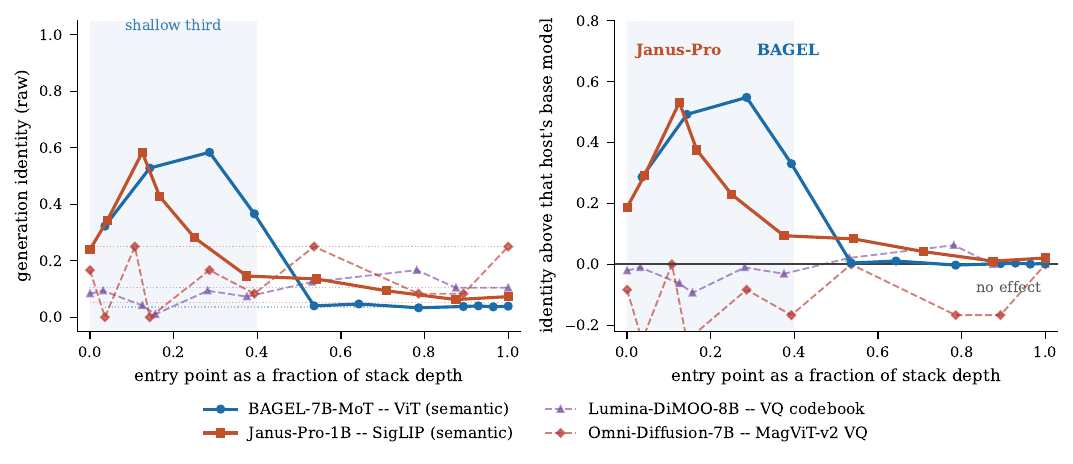}
\caption{The same activation edit on four models, against relative depth so stacks of 24, 28 and 32
blocks are comparable. Curves are grouped by what the understanding pathway encodes: solid for a
semantic vision encoder, dashed for a reconstruction codebook. Left is raw identity with each
model's own measured base level as a faint dotted line; right
subtracts that level, the only quantity comparable across models whose base levels differ
sevenfold. The shaded band is the shallow third, where both semantic-encoder models sit above their
base models and the other two never leave theirs.}
\label{fig:janus}
\end{figure}

The closed-form patch ports without a training stack, which is what makes replication cheap: read
the address from the understanding pathway at layer $L$, read the name's state at the same layer,
rotate, inject, generate, with every weight frozen. We ported it to three further models chosen to
span backbone family, scale and generation mechanism: Janus-Pro-1B \citep{januspro}, autoregressive
over discrete tokens; Lumina-DiMOO-8B \citep{luminadimoo}, discrete diffusion; and
Omni-Diffusion-7B \citep{omnidiffusion}, masked discrete diffusion. Each model is swept over its
whole stack with 12 concepts against a 12-way bank. Base levels are measured by running the
identical code path at $\alpha=0$, not assumed to be the chance line: Omni-Diffusion's base model
scores $0.250$ against a $0.083$ chance line. Janus was
never used to develop the method and each later model was chosen before its curve was seen;
per-model injection modules and probed depths are in Appendix~\ref{app:models}. The result divides
the three: the window reappears on Janus-Pro, while neither Lumina-DiMOO nor Omni-Diffusion shows
an effect at any depth (Table~\ref{tab:arch6}, Figure~\ref{fig:janus}).

\begin{table}[t]
\caption{Peak and base identity for each of the four models, accuracy ($\uparrow$). \emph{Visual peak} is the best site on
the whole depth sweep and \emph{base} is the unmodified model measured on the same generations. The
\emph{word} column rotates the pseudo-name onto a real word's state through the identical code path
and is the instrument check (\S\ref{sec:instrument}). Grouping is by what the understanding
pathway encodes.}
\label{tab:arch6}
\begin{center}
\begin{tabular}{llcccl}
\toprule
model & understanding representation & base & visual peak & word & verdict \\
\midrule
BAGEL-7B-MoT      & ViT, semantic          & 0.036 & 0.584 & ---   & window \\
Janus-Pro-1B      & SigLIP, semantic       & 0.052 & 0.583 & 0.625 & window \\
\midrule
Lumina-DiMOO-8B   & VQ codebook            & 0.104 & 0.167 & 0.656 & null, instr.\ works \\
Omni-Diffusion-7B & MagViT-v2 VQ           & 0.250 & 0.250 & 0.500 & null, instr.\ works \\
\bottomrule
\end{tabular}
\end{center}
\end{table}

\paragraph{Replication on Janus-Pro.} The window is not an artifact of one model. Janus-Pro shares
none of BAGEL's backbone, scale, encoder coupling or generation mechanism, and it reproduces the
shape. The curve rises out of the embeddings and peaks at relative depth $0.12$ with $0.583$,
against its own LoRA-trained ceiling of $0.625$. By half depth it is extinguished ($0.135$ at rel.\
$0.54$ against a base level of $0.052$; Table~\ref{tab:janus}). An
intervention with no gradient steps therefore recovers $93\%$ of what training recovers, against
$80\%$ on BAGEL. What replicates is that a window exists, opens shallow and closes by half depth;
where it peaks does not, since BAGEL is maximized at rel.\ $0.29$ and Janus at rel.\ $0.12$.

\paragraph{Analysis of the two models with no entry-point effect.}\label{sec:instrument}
The two null sweeps peak at $0.167$ against a base level of $0.104$ and at $0.250$ against $0.250$.
On its own neither is yet a finding about the model, because a patch that fires and does nothing is
indistinguishable from a faulty port. Three checks separate the two readings. The patch is applied: every port counts its
own applications, and both models re-run the whole stack at every denoising step, so the edit fires
32 times per image on Lumina-DiMOO and 260 times on Omni-Diffusion. It is not too weak to be detected: it already
reaches $0.62\times$ and $0.51\times$ the state norm against $1.04\times$ on Janus, and raising it
further only degrades the alignment it was meant to install, to centered cosine $0.772$ from
$0.993$ on Lumina. The decisive check runs the same code path with the visual address replaced by a real word's
state at that layer, rotating \emph{brindolt} onto ``elephant''. It is calibrated on Janus, where
the two targets are interchangeable ($0.625$ word against $0.583$ visual). On Lumina-DiMOO and
Omni-Diffusion it moves the model to $6.3\times$ and $2.0\times$ its base level, while the visual
target leaves both at or below base (Table~\ref{tab:archnull}). Both nulls are
therefore about the target rather than the method: the editing machinery works on all four models,
and what these two cannot deliver is a visual address.

The four models split two against two, and the split follows what the understanding pathway
encodes rather than anything the word ``unified'' names. \S\ref{sec:sharing} tests that reading
against the alternatives.

\subsection{Weight sharing versus representation format}\label{sec:sharing}

\paragraph{Weight sharing.} ``Unified'' is a qualitative label, so we made the
axis quantitative: for one real training batch per direction we record the set of parameters each
direction uses and report the parameter-count-weighted Jaccard overlap. The estimator has to
intersect a forward-hook trace with a gradient trace, since either alone is badly wrong. A hook trace
counts modules that are computed and discarded, and a gradient trace counts empty-slice experts,
which scores BAGEL at $0.961$ when the true figure is $0.500$ (Appendix~\ref{app:sharing}). Measured
this way, the model that shares the least transfers the most: BAGEL shares $0.500$ of its backbone
and reaches TransferRate $0.539$ on $G\!\to\!U$ and $0.924$ on $U\!\to\!G$, while Janus-Pro shares
$0.873$ and reaches $0.291$ and $0.473$. These four rates come from the cross-architecture protocol
of 12 concepts against a 12-way bank, not from the 56-concept runs of Table~\ref{tab:main}, so they
are comparable to each other and not to that table. More shared weight is plainly not sufficient for
more transfer. We stop there rather than fitting a trend: two architectures that also differ in scale
cannot separate sharing from anything else, and the split that does hold across all four models is
the one below.

\paragraph{Representation format.} The grouping in Table~\ref{tab:arch6} is
not by backbone family, since BAGEL and Omni-Diffusion share a Qwen2.5-7B lineage and land on
opposite sides, nor by scale, since Janus succeeds at $1.5$B while Lumina-DiMOO fails at $8$B. The
two models where a visual address is usable read their understanding features from a semantic vision
encoder; the two where it is not derive them from a reconstruction objective over a VQ codebook.
Lumina-DiMOO is the most diagnostic case: images and text occupy one vocabulary
and one embedding table, the edit demonstrably lands at centered cosine $0.993$, its derangement
placebo sits at $0.094$, and the address is still unusable, because a VQ index encodes which codebook
entry reproduces a patch and not which object is depicted. Entry point is therefore necessary and not
sufficient. The two directions must also represent concepts in a common semantic format at
that depth, which unified weights do not supply and a semantic understanding encoder does.

\section{A low-cost method for generating images of new concepts}\label{sec:method}

The entry-point rule earns its keep if it improves how concepts are injected in practice.
\S\ref{sec:window} says a binding is usable by both directions when it enters the shared computation
mid-stack, and \S\ref{sec:sar} supplies an objective, Eq.~\ref{eq:anchor}, that writes a binding at
any chosen depth without touching either task's loss. Taken together they prescribe a recipe: apply
the anchoring objective at the mid-stack peak and nowhere else, and skip the generative gradient
entirely. This section evaluates that recipe, and prices every alternative on what it costs the
model's general ability.

\begin{figure}[h]
\centering
\includegraphics[width=0.98\textwidth]{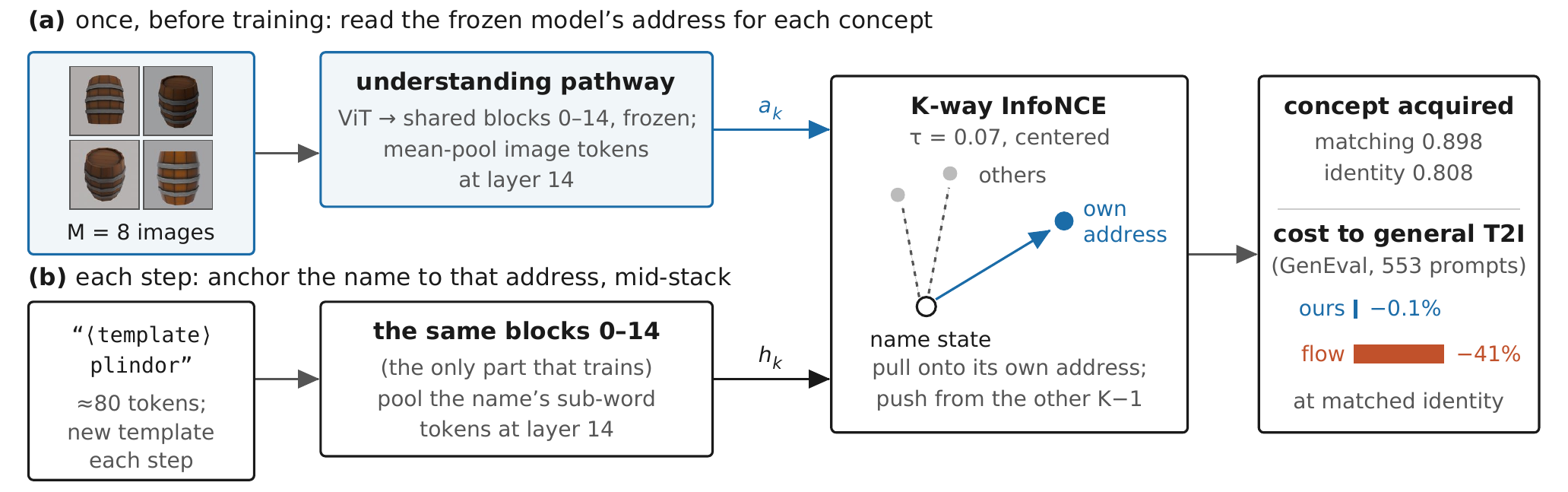}
\caption{Mid-stack Semantic Anchoring. (a) Once, before training, each concept's images are embedded
through the frozen understanding pathway and the shared expert's states are mean-pooled at layer 14
into a frozen visual address $a_k$. (b) At each step, a short prompt containing the name runs to the
same layer, the name's sub-word states are pooled into $h_k$, and a $K$-way InfoNCE pulls $h_k$ onto
$a_k$ and away from the other concepts' addresses. No flow-matching gradient is ever computed, which
is where the generative route's $41\%$ GenEval cost originates (Table~\ref{tab:geneval}); anchoring
costs $0.1\%$ and acquires the concept for both directions (Table~\ref{tab:main}).}
\label{fig:method}
\end{figure}

\paragraph{Mid-stack Semantic Anchoring.} The recipe instantiates Eq.~\ref{eq:anchor} at layer
$L=14$ (Figure~\ref{fig:method}). Before training, embed $M=8$ images of each concept through the
model's own understanding pathway and mean-pool the shared expert's hidden states at layer $L$;
freeze those $K$ addresses $a_k$. At each step, run $K$ short prompts ``\emph{template} \{name\}''
($\approx$80 tokens in total), pool over each name's own sub-word tokens at the same layer to get
$h_k$, and take one step on the anchoring loss. The template is resampled each step. The recipe
uses no supervision either task lacks, the same images and the same names, and it adds no module or
token beyond the LoRA adapter every baseline here also trains. One $\approx$80-token forward and
backward per step is its whole cost. It computes no generative gradient, which turns out to be the
source of the damage.

\paragraph{Concept acquisition without a flow-matching gradient.} Run the anchor alone for 480 steps with the flow-matching
objective switched off: over 56 concepts the model reaches $0.898$ \ci{0.84}{0.94} name matching and
$0.808$ \ci{0.72}{0.89} identity, against $0.520$ and $0.653$ for a flow-matching baseline at
comparable wall-clock (Table~\ref{tab:main}). Paired within group, the matching advantage is
$+0.379$ \ci{+0.29}{+0.47} with $7/7$ agreeing; the identity advantage is $+0.155$
\ci{+0.00}{+0.29} with $6/7$ and a lower bound on zero. Anchoring alone is far better on the task it was never given, and level on the
task the flow-matching objective exists to serve. 

\begin{table}[t]
\caption{General text-to-image ability retained after each recipe. GenEval \citep{geneval} prompts, all
553, two images each, identical seeds, all columns accuracy ($\uparrow$). We substitute an OWLv2
detector \citep{owlv2} for the official Mask2Former, so absolute scores are not comparable to
published GenEval numbers and every claim here is a $\Delta$ against the same base model under the
same detector. \emph{Steps} counts the optimizer steps
the recipe takes, and \emph{flow steps} how many of them compute a flow-matching gradient.}
\label{tab:geneval}
\begin{center}
\begin{tabular}{lcccccc}
\toprule
condition & steps & flow steps & overall & $\Delta$ & rel. & identity \\
\midrule
base (unmodified)                       & ---  & ---  & 0.681 & ---      & ---      & 0.036 \\
\midrule
activation patch @ layer 3              & 0    & 0    & 0.681 & $0.000$  & $0.0\%$  & \bt{0.930} \\
anchoring @ layer 14, alone             & 480  & 0    & 0.680 & $-0.001$ & $0.1\%$  & \bt{0.930} \\
U-side, shared MLP only                 & 1600 & 0    & 0.680 & $-0.001$ & $0.1\%$  & 0.445 \\
\midrule
U-side then a generation phase          & 1760 & 160  & 0.680 & $-0.001$ & $0.1\%$  & 0.891 \\
anchoring @ layer 14 $+$ flow matching  & 480  & 480  & 0.638 & $-0.042$ & $6.2\%$  & 0.953 \\
joint gen$+$und                         & 640  & 640  & 0.692 & $+0.011$ & $-1.6\%$ & 0.820 \\
generative default                      & 960  & 960  & 0.401 & $-0.279$ & $41.0\%$ & 0.922 \\
\bottomrule
\end{tabular}
\end{center}
\end{table}

\paragraph{Attribution of the GenEval loss.} Table~\ref{tab:geneval} quantifies the cost of each
recipe on 553 held-out prompts containing no pseudo-name. Cost and benefit come from the same runs,
so its last column is measured on the same group of eight as its damage columns. The two
routes are priced where they acquire the concept equally well, identity $0.930$ against $0.922$,
rather than at equal steps, which 480 steps of flow matching do not reach ($0.469$). The standard
generative route costs 28 points of the prompt suite, $41\%$ of the model's measured general
text-to-image ability. The loss is not diffuse: it concentrates in the compositional categories,
two-object at $-0.485$ and spatial position at $-0.370$, while single-object prompts survive at
$-0.150$. A model fine-tuned on eight isolated objects largely forgets how to put two things in one
picture and where. Across the table every point of damage coincides with the presence of a
flow-matching gradient: the two methods that never compute one cost $0.1\%$ each, and a 160-step
generation phase costs $0.1\%$, a 480-step phase $6.2\%$, a 960-step phase $41\%$. Anchoring alone also places the concept in unseen scenes at identity
$0.963$ with CLIP-T $0.230$, above the base model's $0.222$ where the generative route reaches
$0.191$, so composition survives and not merely the aggregate score.

\paragraph{Identity against a same-category sibling.} A 24-way retrieval over single assets from
distinct categories could in principle be won by drawing a generic member of the right category.
Rebuilding three groups from same-category sibling pairs removes that shortcut, and every trained
condition still tells its asset from its own sibling at $0.69$ to $0.91$ against the base model's $0.484$. The residual shortcut lands exactly where the mechanism predicts: a sibling in the
bank costs flow matching $+0.182$ \ci{+0.09}{+0.29} and the two address-aligning interventions
$+0.055$ and $+0.016$, both intervals containing zero. A pixel-level target can be partly satisfied
by a category-typical member; an address-alignment target cannot, because the address belongs to one
asset. Appendix~\ref{app:instance} gives the full study and the resulting correction to identity
levels elsewhere, roughly $0.18$ for flow matching and $0.05$ for anchoring.

\section{Discussion}\label{sec:discussion}
\paragraph{Scope of the claim.} The defensible conclusion is that the usability of an injected concept binding depends strongly on where it enters downstream computation shared across tasks, and that the two transfer directions must represent concepts in a common semantic format at that point. Three observations support this view: the embedding layer leaves the greatest amount of downstream computation, yet injection there fails; attention-only injection traverses the entire network, yet does not transfer to the other task; and the effective entry-point windows differ in width between the two directions. Extending the common-semantic-format requirement to the training target suggests that the operative variable should be whether the generation target forces information through the shared semantic pathway.

\paragraph{Limitations.} Five aspects of the design
bound what is claimed here, and each marks a concrete next experiment. (i) ``Computation after the entry point'' is operationalized as layer count, and the
block-deletion experiment built to refine it separates the two accounts only in part, because
deletion is not equally costly on the two sides of a site: two blocks immediately before the entry
point cost fifteen times more general prompt fidelity than two immediately after, so at comparable
damage both arms sit at the base level (Appendix~\ref{app:downstream}). (ii) The window is measured across four architectures, an observational series in which
encoder type covaries with backbone, scale and generation mechanism, so training one architecture
twice, once with a semantic and once with a reconstruction encoder, would turn the regularity into
a controlled result. (iii) Identity is retrieval under two automatic encoders that agree closely
with each other ($r = 0.997$ across the depth sweep) rather than under human judgement, which
leaves the absolute levels open to calibration by a human study. (iv) BAGEL's private generation
expert is too weak to hold a concept on its own, so whether stronger private branches silo
knowledge is untested. (v) Runs use eight concepts each by design, since
concept count is task difficulty, which leaves open how the window behaves when a model is asked to
hold hundreds of bindings at once.

\paragraph{Implications and future research.} These findings reframe concept learning in unified multimodal models as a problem of routing and representation, rather than solely one of parameter sharing or training scale. For model design, they suggest that new concepts should be introduced where understanding and generation still share both a compatible semantic format and sufficient downstream computation, which gives a principled basis for choosing adaptation sites instead of treating layers as interchangeable. For future research, the entry-point window offers a testable diagnostic for comparing architectures, objectives, and private-versus-shared pathways, while the semantic-format hypothesis motivates controlled studies that vary the generation target without changing the data. More broadly, identifying where knowledge becomes usable across tasks could guide more efficient multimodal adaptation, reduce unnecessary parameter updates, and distinguish models that merely store a concept from those that can deploy it across modalities and tasks.



\section{Conclusion}\label{sec:conclusion}

In this paper, we asked whether a unified multimodal model can move a newly bound concept between understanding and
generation, and separated that architectural question from the data question by binding a novel
entity through exactly one direction and measuring the other. The channel is real both ways, but the
directions differ in kind: generation training installs a name the model can match and not produce.
What governs usability is where the binding enters. Alignment predicts export without causing
it, while the same objective's closed-form edit over activations works at layer 7 of 28 and is
indistinguishable from the base model by layer 14, and carried by weights peaks at layers 10--14. Across four models the window appears
only where the understanding pathway is a semantic vision encoder, so unified weights are not
sufficient. Acting on that, we anchor a name onto the model's own visual address at layer 14, with no
generative gradient anywhere in the objective. This acquires 56 concepts at $0.898$ name matching
and $0.808$ identity, for a $0.1\%$ relative loss of general text-to-image ability where the
generative route costs $41\%$. Entry point is a design variable that unified models already have and that their
training recipes do not use. This paper leaves behind the means to use it: a contamination-free
measurement of cross-task usability, an intervention that isolates depth from optimization by
taking no gradient steps at all, and a condition on the host architecture that says where the
variable takes effect.

\bibliographystyle{tmlr}
\bibliography{main}

\appendix

\section{Discovery and confirmation protocol}\label{app:protocol}

Every contrast in this paper was found on group g0. The other six groups were rendered,
trained and evaluated afterwards against a fixed contrast list and are reported as a held-out
confirmation set (Table~\ref{tab:confirm}). All four primary contrasts replicated across all six
confirmation groups. We report both discovery and confirmation results, and where a magnitude
carries an argument we use the confirmation column.

\begin{table}[h]
\caption{Discovery versus confirmation. g0 is where each contrast was found; g1--g6 were run
afterwards against a fixed list. The final column reports the discovery-to-confirmation ratio.}
\label{tab:confirm}
\begin{center}
\begin{tabular}{lcccc}
\toprule
paired contrast & discovery (g0) & confirmation (g1--g6) & pooled & ratio \\
\midrule
alignment @14 vs @ readout, identity   & $+0.906$ & $+0.672$ \ci{+0.59}{+0.74} 6/6 & $+0.705$ & $1.35\times$ \\
alignment @14 vs @ readout, matching   & $+0.637$ & $+0.554$ \ci{+0.50}{+0.60} 6/6 & $+0.566$ & $1.15\times$ \\
alignment alone vs G-inject, matching  & $+0.362$ & $+0.381$ \ci{+0.27}{+0.49} 6/6 & $+0.379$ & $0.95\times$ \\
shared MLP vs shared attention, export & $+0.516$ & $+0.219$ \ci{+0.14}{+0.31} 6/6 & $+0.261$ & $2.36\times$ \\
\midrule
\emph{alignment alone vs G-inject, identity} & $+0.273$ & $+0.135$ \ci{-0.04}{+0.29} 5/6 & $+0.155$ & $2.02\times$ \\
\emph{G-inject into shared MLP vs default}      & $+0.148$ & $+0.049$ \ci{-0.06}{+0.17} 2/6 & $+0.064$ & $3.00\times$ \\
\bottomrule
\end{tabular}
\end{center}
\end{table}

The SAR correlation, by family.
After removing four duplicate configurations, Table~\ref{tab:sar} reports within-family
correlations, correlations pooled over all remaining families, and leave-one-family-out MAE.

\begin{table}[h]
\caption{SAR versus export with the dependence structure respected. $\rho$ within is the correlation
inside one sweep; $\rho$ without is the pooled correlation with that sweep deleted; MAE is the error
of predicting that sweep's configurations from a fit that never saw them, in accuracy units.}
\label{tab:sar}
\begin{center}
\begin{tabular}{lcccc}
\toprule
configuration family & $n$ & $\rho$ within & $\rho$ without & MAE predicting it \\
\midrule
capacity allocation & 16 & $+0.46$ & $+0.78$ & 0.200 \\
placement / main    & 10 & $+0.60$ & $+0.59$ & 0.137 \\
recipe              &  6 & $+0.71$ & $+0.67$ & 0.194 \\
substrate           &  4 & $+0.80$ & $+0.66$ & 0.122 \\
\midrule
\textbf{pooled}     & \textbf{36} & \multicolumn{3}{l}{$\boldsymbol{+0.68}$, family-clustered bootstrap $[+0.37,+0.88]$} \\
\bottomrule
\end{tabular}
\end{center}
\end{table}

Direct-task accuracy and SAR were evaluated for each configuration.

\section{Probe and evaluation details}\label{app:probes}

\paragraph{The context-free production probe is scored, not sampled.} No text is generated and no
decoding hyper-parameter exists to tune. For each held-out image and each candidate pseudo-name in
the group we run one forward pass over the fixed string ``$\langle$image$\rangle$ This is a
\{name\}'' and read the model's log-probability of the name's sub-word sequence; the prediction is
the argmax over candidates. There is no temperature, no top-$k$/top-$p$, and no sampling seed, so
the probe is deterministic given the image set.

We corrected unconditional name probability with pointwise mutual information and fixed
first-sub-word-only scoring before running the evaluation. We also recorded full-name scores for all
conditions in Table~\ref{tab:matching}.

\begin{figure}[h]
\centering
\includegraphics[width=0.8\textwidth]{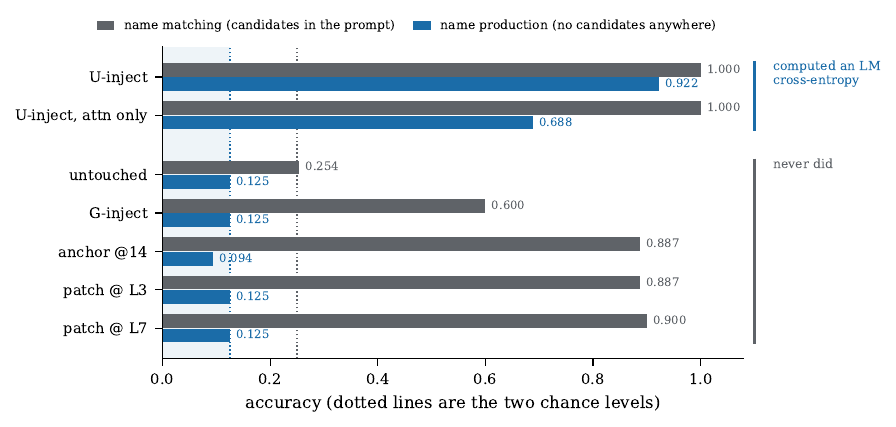}
\caption{The same conditions scored with the candidate names in the prompt and with them
removed, grouped by evaluation condition. The shaded region is production chance.}
\label{fig:matching}
\end{figure}

Figure~\ref{fig:matching} plots the same split.

Per-concept identity distribution.
We computed identity per concept over 16 generations at layer 7. Of 56 concepts, 17 score
$1.000$, 6 score $0.000$, and the remainder lie between; uncertainty is reported with the
pre-specified group-level bootstrap.

Encoder agreement at site and concept levels.
We computed agreement between the two encoders across twelve sites and across 56 individual
concepts at layer 7. The site-level correlation is $r=0.997$; the concept-level correlation is
$r=0.821$, with mean absolute difference $0.165$ and the same above-floor verdict on $46/56$.

\paragraph{Notation for the ``shared$+$private'' rows.} In the U-inject conditions the adapter is
instantiated on both experts, but the understanding pass routes through the shared and
understanding-side experts only, so the generation-side private expert receives no gradient and is
bit-identical to its initialization at the end of training. The row label describes where parameters
were 
placed, not where they were updated. At evaluation both branches are active in every
condition and the routing is the model's own; we never disable an expert at test time.

\section{The closed-form activation edit}\label{app:closedform}

This appendix specifies the objective, intervention, closed-form edit, evaluated variants,
residual measurements, and stability protocol used for the activation arm.

\paragraph{The objective.} Fix a layer $L$ and a group of $K$ concepts. For concept $k$, let
$v_k \in \mathbb{R}^D$ be its 
visual address: the shared expert's layer-$L$ hidden state
mean-pooled over the image tokens of $M=8$ understanding-side images, averaged over images. Let
$t_k \in \mathbb{R}^D$ be its 
name state: the layer-$L$ hidden state of a short prompt
ending in the pseudo-name, mean-pooled over that name's own sub-word token positions. Write
$\bar{t} = \tfrac1K\sum_j t_j$ and $t_k^{c} = t_k - \bar{t}$, and likewise $v_k^c$. Semantic
Anchoring minimizes the $K$-way InfoNCE
\begin{equation}
\mathcal{L} \;=\; -\frac{1}{K}\sum_{k=1}^{K}
\log \frac{\exp\!\big(\cos(t_k^{c}, v_k^{c})/\tau\big)}
          {\sum_{j=1}^{K}\exp\!\big(\cos(t_k^{c}, v_j^{c})/\tau\big)},
\qquad \tau = 0.07 .
\label{eq:infonce}
\end{equation}
Both sides are centered across the group before the cosine. That is the whole reason a closed form
exists: centering removes the constant text-versus-image offset, so the objective constrains only the
direction of each concept's deviation from its group mean, and says nothing at all about the
mean or about any deviation magnitude.

\paragraph{The intervention.} Freeze every weight. Introduce one vector $\delta_k \in \mathbb{R}^D$
per concept, added to the residual stream at layer $L$ at exactly the token positions where
concept $k$'s name appears, so the patched state is $t_k' = t_k + \delta_k$. Positions are recovered
at layer 0 by matching the name's exact sub-word embedding sequence, so training and evaluation
locate them identically and the intervention is self-contained.

\begin{figure}[h]
\centering
\includegraphics[width=\textwidth]{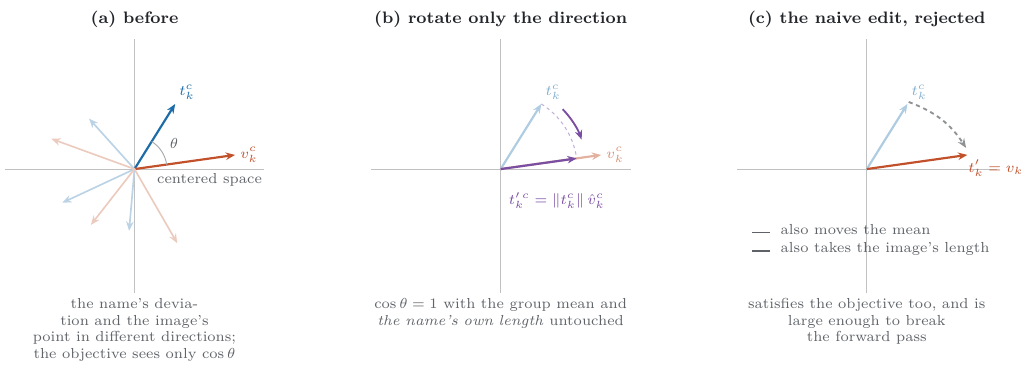}
\caption{The edit in the centered space where the objective is evaluated. The reported
experiments use the direction-only edit in panels (a)--(b); panel (c) shows the separately evaluated
naive edit.}
\label{fig:edit}
\end{figure}

\paragraph{The solution.} Hold the center fixed at $\bar{t}$ and set
\begin{equation}
\boxed{\;\delta_k \;=\; \bar{t} \;+\; \lVert t_k^{c}\rVert \cdot
\frac{v_k^{c}}{\lVert v_k^{c}\rVert} \;-\; t_k\;}
\qquad\Longrightarrow\qquad
t_k' - \bar{t} \;=\; \lVert t_k^{c}\rVert\,\hat{v}_k^{c},
\label{eq:closedform}
\end{equation}
which is parallel to $v_k^c$ and therefore attains $\cos = 1$ on every diagonal term of
Eq.~\ref{eq:infonce} simultaneously. That is the global maximum of the alignment term, at every
depth, with no gradient step. It is not in general the minimizer of Eq.~\ref{eq:infonce} itself: the
negative terms $\cos(t_k^{c'}, v_j^c)$ are then fixed at $\cos(v_k^c, v_j^c)$ by the addresses
themselves and are not driven down, and holding the center at the pre-patch $\bar t$ is a constraint
rather than an identity, since the patched states' own mean is $\bar t + \tfrac1K\sum_k \lVert
t_k^c\rVert \hat v_k^c$, which need not equal $\bar t$. Both approximations are what the residual
measurements below quantify, and neither is assumed away. Two constraints are imposed deliberately and neither is required by the objective:
the patched state keeps the group mean $\bar t$, and it keeps 
its own deviation norm
$\lVert t_k^c \rVert$ rather than the image's. Only the direction rotates. Among all $\delta_k$
achieving $\cos=1$ this is the unique minimizer of $\lVert \delta_k \rVert$ subject to preserving
both, which is what we mean by calling it minimal (Figure~\ref{fig:edit}).

Evaluated edit variants.
We evaluated both $t_k'=v_k$ and the minimal edit of Eq.~\ref{eq:closedform}. All reported
activation-arm results use the minimal edit. At layer 14 it has $\lVert\delta\rVert=640$ against a
name-state norm of $1602$, a relative size of $0.40$.

Residual measurements.
We measured re-centring and template-averaging residuals for all \NSOLVE{} closed-form solves
in the seven-group sweep. The post-patch anchoring loss is \RESMEAN{} on average against chance
$\log 8=2.0794$ and reaches \RESMAX{} at its maximum; the corresponding worst-case reduction from
chance toward zero is \RESWORST{}.

The single-group table, including the flow-matching arm.
Table~\ref{tab:entry} reports the original eight-concept measurement for anchoring-only
activation edits, anchoring-only weight edits, and weight edits trained with anchoring plus flow
matching. Seven-group levels are reported in Table~\ref{tab:entry7g}.

\begin{table}[h]
\caption{The depth sweep on one group of eight, three arms, accuracy ($\uparrow$). All optimize the same anchoring
objective at the same site. The first two differ 
only in medium; the third adds flow matching.
The base model scores $0.036$ on identity and $0.254$ on matching.}
\label{tab:entry}
\begin{center}
\begin{tabular}{lcccccc}
\toprule
& \multicolumn{3}{c}{generation identity} & \multicolumn{3}{c}{name matching} \\
\cmidrule(lr){2-4}\cmidrule(lr){5-7}
entry point & activations & weights & \emph{weights} & activations & weights & \emph{weights} \\
            & anchor only & anchor only & \emph{+ flow} & anchor only & anchor only & \emph{+ flow} \\
\midrule
layer 3      & \bt{0.930} & \bt{0.945} & \emph{0.949} & 0.887 & 0.900 & \emph{0.900} \\
layer 7      & 0.828 & 0.914 & \emph{0.984} & 0.900 & 0.925 & \emph{0.950} \\
layer 10     & 0.469 & 0.938 & --- & 0.862 & 0.962 & --- \\
layer 14     & 0.016 & \bt{0.930} & \emph{0.984} & 0.500 & 0.887 & \emph{0.921} \\
layer 21     & 0.023 & 0.633 & \emph{0.988} & 0.113 & 0.787 & \emph{0.894} \\
layer 24     & 0.023 & 0.406 & \emph{0.965} & 0.138 & 0.775 & \emph{0.919} \\
layer 26     & 0.023 & 0.203 & \emph{0.977} & 0.138 & 0.663 & \emph{0.856} \\
layer 27     & 0.016 & 0.055 & \emph{0.625} & 0.138 & 0.350 & \emph{0.662} \\
after the norm & 0.016 & 0.023 & --- & 0.138 & 0.250 & --- \\
\bottomrule
\end{tabular}
\end{center}
\end{table}

Composition and stability.
We applied the edit at one site at a time. Across depths we measured
$\lVert\delta\rVert/\lVert t\rVert=0.56$ at layer 0, $0.40$ at layer 14, and $0.38$ at layer 24;
the magnitude sweep therefore reports this ratio and varies $\alpha$.

\section{Capacity allocation}\label{app:capacity}

Holding the adapter budget fixed at 80.74M trainable parameters and sliding it between BAGEL's two
experts asks a question placement ablations cannot: not whether an adapter 
confined to the
private expert can transfer, but what an objective 
free to use either does with a private
store when one is available.

Table~\ref{tab:capacity} reports the full constant-budget sweep.

\begin{table}[h]
\caption{Constant-budget capacity allocation, group of eight, accuracy ($\uparrow$); the italicised all-private row is
the measured base level of this sweep, $0.138$ for matching and $0.016$ for retrieval, and is what
the last column subtracts; analytic chance is $0.250$ and $0.042$. Trainable parameters are
identical in every row. Export efficiency is (cross $-$ base)/(direct $-$ base), blank when the
direct task is within $0.15$ of that base.}
\label{tab:capacity}
\begin{center}
\begin{tabular}{llccccc}
\toprule
direction & $r_{\text{und}}$ & $r_{\text{gen}}$ & private share & direct & cross & export eff. \\
\midrule
G-inject & 32 & 0  & 0.00 & 0.773 & 0.575 & 0.58 \\
G-inject & 24 & 8  & 0.25 & \bt{0.883} & \bt{0.700} & 0.65 \\
G-inject & 16 & 16 & 0.50 & 0.781 & 0.525 & 0.51 \\
G-inject & 8  & 24 & 0.75 & 0.531 & 0.412 & 0.53 \\
G-inject & 0  & 32 & 1.00 & 0.102 & 0.138 & --- \\
\midrule
U-inject & 32 & 0  & 0.00 & \bt{1.000} & 0.531 & 0.60 \\
U-inject & 24 & 8  & 0.25 & \bt{1.000} & 0.445 & 0.50 \\
U-inject & 16 & 16 & 0.50 & \bt{1.000} & 0.398 & 0.44 \\
U-inject & 8  & 24 & 0.75 & \bt{1.000} & 0.242 & 0.26 \\
U-inject & 0  & 32 & 1.00 & \emph{0.138} & \emph{0.016} & --- \\
\bottomrule
\end{tabular}
\end{center}
\end{table}

\begin{figure}[h]
\centering
\includegraphics[width=0.72\textwidth]{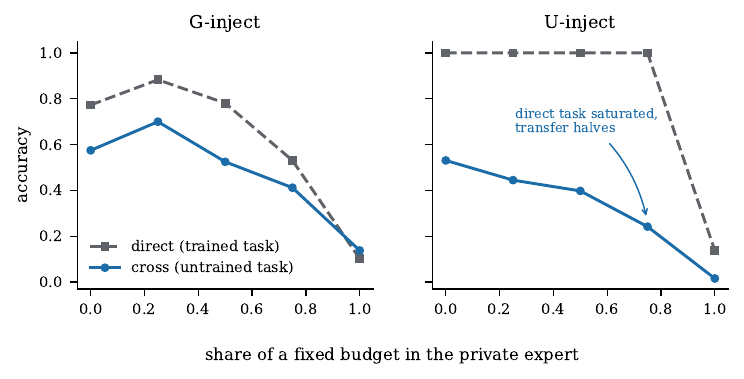}
\caption{A fixed adapter budget slid between the shared and private experts. In the U-inject panel
the trained task is saturated at every allocation while transfer halves, so the shared capacity
transfer needs is strictly greater than the capacity the task itself needs.}
\label{fig:capacity}
\end{figure}

Cross-task transfer falls monotonically as budget moves to the private expert
(Figure~\ref{fig:capacity}), in both directions and
robustly to concept resampling (Spearman $-0.90$, bootstrap $[-1.00,-0.30]$ for G-inject; $-1.00$,
$[-1.00,-0.60]$ for U-inject). At the fully private G-inject endpoint, transfer is exactly at chance.

Matched-rank private-capacity control.
We held shared rank fixed and added capacity on the private side at three matched shared
ranks. Cross-task accuracy changed by $-0.01$ on average; the direct score at the fully private
endpoint is $0.117$.

\section{The block-deletion experiment}\label{app:downstream}

Every measurement in \S\ref{sec:window} moves the entry point and reads depth off the layer index, so
depth and downstream computation move together and the thesis' own independent variable is never
manipulated on its own. We therefore hold the site fixed at layer 7 and remove computation instead,
making $k$ consecutive decoder blocks pass their residual stream through unchanged. Two windows of
equal size are compared: the $k$ blocks immediately 
after the site, which the injected state
must traverse, and the $k$ blocks immediately 
before it, which it never does.

Two design points matter. First, general damage is measured directly rather than assumed matched: a
separate run over ordinary prompts containing no pseudo-name reports CLIP text--image agreement under
the same ablation. Second, 
the patch is re-solved under each ablation, since a patch computed
on the intact model is the wrong patch for a network missing four of its blocks.

\begin{table}[h]
\caption{Deleting computation with the entry point held fixed at layer 7, identity
($\uparrow$). Every arm's patch is re-solved under its own ablation. $\Delta$CLIP-T is measured on
ordinary prompts containing no pseudo-name.}
\label{tab:downstream}
\begin{center}
\begin{tabular}{llccc}
\toprule
blocks deleted & where & which blocks & concept identity & $\Delta$CLIP-T (general ability) \\
\midrule
0 & ---                & ---                  & 0.828 \ci{0.66}{0.97} & $\pm0.0000$ \\
\midrule
2 & after the site     & 8, 9                 & \bt{0.742} \ci{0.56}{0.91} & $-0.0065$ \\
4 & after the site     & 8--11                & 0.102 \ci{0.02}{0.20} & $-0.0832$ \\
6 & after the site     & 8--13                & 0.016 \ci{0.00}{0.05} & $-0.1084$ \\
\midrule
2 & before the site    & 5, 6                 & 0.000 \ci{0.00}{0.00} & $-0.1018$ \\
4 & before the site    & 3--6                 & 0.000 \ci{0.00}{0.00} & $-0.0909$ \\
6 & before the site    & 1--6                 & 0.000 \ci{0.00}{0.00} & $-0.0879$ \\
\bottomrule
\end{tabular}
\end{center}
\end{table}

Measured block-deletion controls.
We deleted $k\in\{2,4,6\}$ consecutive blocks immediately before or after a fixed layer-7
entry point, re-solved the patch for every ablation, and measured both concept identity and
$\Delta$CLIP-T on ordinary prompts. Table~\ref{tab:downstream} reports both measurements; this
ablation is not used for causal attribution.

\section{Parameter-sharing estimators}\label{app:sharing}

For one real training batch per direction we record the set of parameters each direction 
uses
and report the parameter-count-weighted Jaccard overlap of the two sets. The backbone scope restricts
this to the shared transformer stack and is the figure quoted in \S\ref{sec:sharing}; the whole-model
scope additionally charges each direction for its private codecs and heads.

Intersection estimator.
We counted a parameter as used only when its module fired on a non-empty input and remained
gradient-connected. Parameter names were canonicalized by tensor identity before computing the
parameter-count-weighted Jaccard overlap. Hook-only, gradient-only, and intersection estimates are
reported in Table~\ref{tab:sharing}.

\begin{table}[h]
\caption{Measured cross-task parameter sharing for the two trained models under hook-only,
gradient-only, and intersection estimators. Bold marks the reported intersection estimate.}
\label{tab:sharing}
\begin{center}
\begin{tabular}{lccccc}
\toprule
 & & \multicolumn{3}{c}{backbone sharing by estimator} & \\
\cmidrule(lr){3-5}
architecture & params & hook only & grad only & \bt{intersection} & whole model \\
\midrule
BAGEL-7B-MoT \citep{bagel}   & 14.1B & 0.500 & \bt{0.961} & \bt{0.500} & 0.484 \\
Janus-Pro-1B \citep{januspro} & 1.5B & 0.873 & 0.873 & \bt{0.873} & 0.720 \\
\bottomrule
\end{tabular}
\end{center}
\end{table}

\section{Per-model replication details}\label{app:models}

\S\ref{sec:format} groups four models by their understanding representation. Table~\ref{tab:hostdetail}
gives the evidence behind that grouping in full, so a reader can check the classification rather than
take it.

\begin{table}[h]
\caption{Every model in the four-architecture study. 
Injection module is the exact module list
the forward hook was registered on; the patch always fires in the language backbone's residual
stream, never in a vision tower or a decoder, so ``depth'' means the same thing on all four. Sites are
absolute block indices; $-2$ is the input embedding and $-1$ the last block's output after the final
norm.}
\label{tab:hostdetail}
\begin{center}
\footnotesize
\begin{tabular}{@{}llllc@{}}
\toprule
model & backbone & understanding pathway & generation pathway & blocks \\
\midrule
BAGEL-7B-MoT      & Qwen2.5-7B (MoT)       & ViT, semantic          & rectified flow, VAE latent & 28 \\
Janus-Pro-1B      & Llama, 1.5B            & SigLIP, semantic       & LlamaGen VQ                & 24 \\
Omni-Diffusion-7B & Dream-7B (Qwen2.5-7B)  & MagViT-v2 VQ           & masked discrete diffusion  & 28 \\
Lumina-DiMOO-8B   & LLaDA-8B               & VQ codebook            & discrete diffusion         & 32 \\
\bottomrule
\end{tabular}
\end{center}
\end{table}

\begin{center}\footnotesize
\begin{tabular}{@{}lll@{}}
\toprule
model & injection module (hooked) & sites probed / concepts / bank / $n_{\text{gen}}$ \\
\midrule
BAGEL & \texttt{language\_model.model.layers} & $\{-2,0,3,7,10,14,17,21,24,25,26,27,-1\}$ / 8 / 24 / 16 \\
Janus & \texttt{language\_model.model.layers} & $\{-2,0,2,3,5,8,12,16,20,23,-1\}$ / 12 / 12 / 8 \\
Omni-Diffusion & \texttt{model.layers} & $\{-2,0,2,3,7,10,14,21,24,-1\}$ / 12 / 12 / 8 \\
Lumina-DiMOO & \texttt{model.transformer.blocks} & $\{-2,0,3,4,8,11,16,24,27,-1\}$ / 12 / 12 / 8 \\
\bottomrule
\end{tabular}
\end{center}

For all four models we recorded the understanding pathway, generation pathway, backbone,
scale, hooked module list, probed sites, concept count, bank size, and generation count before
comparing their depth curves.

Full depth curves for the two 2026 models.
We measured every probed site for both models. Lumina-DiMOO spans $0.010$ to $0.167$ against
a floor of $0.104$, and Omni-Diffusion spans $0.000$ to $0.250$ against a floor of $0.250$.
Per-site values are released with the code. Scaling the edit does not change either null: on
Lumina-DiMOO at layer 4, $\alpha \in \{1,2,5\}$ gives $0.010$, $0.052$ and $0.146$ against a base
level of $0.104$, the last at $3.12\times$ the state norm; on Omni-Diffusion at layer 3 the same
sweep gives $0.000$, $0.083$ and $0.083$ against a base level of $0.250$, the last at $2.57\times$.
Table~\ref{tab:archnull} gives the per-model word-target instrument check at each model's best site.

\begin{table}[h]
\caption{The entry-point window on Janus-Pro against BAGEL at the nearest relative depth, identity
($\uparrow$). The intervention is identical in closed form, with all weights frozen and no gradient step; only the model changes. BAGEL figures are the seven-group values of
Table~\ref{tab:entry7g} except where marked.}
\label{tab:janus}
\begin{center}
\begin{tabular}{lccc}
\toprule
site & relative depth & Janus-Pro-1B & BAGEL at the nearest relative depth \\
\midrule
input embeddings & 0.00 & 0.240 & \emph{0.023} (embeddings, 1 group) \\
layer 0          & 0.04 & 0.344 & 0.323 (layer 0) \\
\textbf{layer 2} & \textbf{0.12} & \bt{0.583} & 0.528 (layer 3, rel.\ 0.14) \\
layer 3          & 0.17 & 0.427 & 0.528 (layer 3) \\
layer 5          & 0.25 & 0.281 & \bt{0.584} (layer 7, rel.\ 0.29) \\
layer 8          & 0.38 & 0.146 & 0.366 (layer 10) \\
layer 12         & 0.54 & 0.135 & 0.040 (layer 14) \\
layer 16         & 0.71 & 0.094 & 0.047 (layer 17) \\
layer 20         & 0.88 & 0.062 & 0.038 (layer 24) \\
last block       & 1.00 & 0.073 & 0.039 (last block) \\
\midrule
\emph{layer 3, deranged addresses} & 0.17 & \emph{0.052} & \emph{0.000} \\
\emph{layer 5, deranged addresses} & 0.25 & \emph{0.115} & --- \\
\emph{base model}             & ---  & \emph{0.052} & \emph{0.036} \\
\emph{LoRA trained on generation}  & ---  & \emph{0.625} & \emph{0.724} \\
\bottomrule
\end{tabular}
\end{center}
\end{table}

\begin{table}[h]
\caption{The word-target instrument check of \S\ref{sec:instrument}, identity
($\uparrow$). Each row rotates a pseudo-name onto a target at the model's best site through one code
path; only the target differs. Percentages use each model's measured floor and available trained
ceiling. ``$<$floor'' marks a score below the measured floor.}
\label{tab:archnull}
\begin{center}
\begin{tabular}{lccccc}
\toprule
model & site & visual address & real-word target & floor & trained ceiling \\
\midrule
BAGEL-7B-MoT & layer 7 & \bt{0.584} (80\%) & --- & 0.036 & 0.724 \\
Janus-Pro-1B & layer 2 & \bt{0.583} (93\%) & \bt{0.625} (100\%) & 0.052 & 0.625 \\
Lumina-DiMOO-8B & layer 4 & 0.010 ($<$floor) & \bt{0.656} (6.3$\times$ floor) & 0.104 & --- \\
Omni-Diffusion-7B & layer 3 & 0.000 ($<$floor) & \bt{0.500} (2.0$\times$ floor) & 0.250 & --- \\
\bottomrule
\end{tabular}
\end{center}
\end{table}

\section{Instance versus category}\label{app:instance}

We evaluated instance identity separately from category identity using single assets drawn
from Objaverse and same-category sibling controls.

\begin{figure}[h]
\centering
\includegraphics[width=0.66\textwidth]{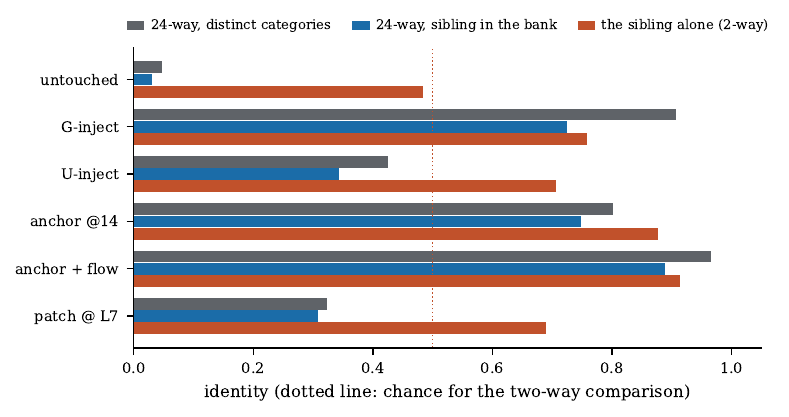}
\caption{The same generations scored three ways. Gray is the bank the rest of the paper uses, where
no distractor shares the target's category; blue puts a same-category sibling in the bank; orange
compares the target against that sibling alone, where the category carries no information at all.}
\label{fig:finegrained}
\end{figure}

We used 12 same-category sibling pairs, 24 concepts, formed into three groups of eight from four pairs each,
and scored the same generations with a 24-way coarse bank, a 24-way fine bank, a two-way sibling
comparison, and CLIP's vision tower. On the understanding side, the two same-category names were
compared directly. Figure~\ref{fig:finegrained} shows the three scorings of the same generations,
and Table~\ref{tab:finegrained} reports every condition.

\begin{table}[h]
\caption{Instance identity, three fine-grained groups (24 concepts in 12 same-category pairs),
accuracy ($\uparrow$). 
Coarse and 
fine are the same 24-wide width scored on the
same generated images and differ only in whether one distractor shares the target's category.
Paired drop is coarse minus fine, differenced per concept. Chance is $0.042$ for the 24-way
columns and $0.500$ for the two-way ones.}
\label{tab:finegrained}
\begin{center}
\begin{tabular}{lccccc}
\toprule
 & \multicolumn{3}{c}{generation identity} & \multicolumn{2}{c}{understanding} \\
\cmidrule(lr){2-4}\cmidrule(lr){5-6}
condition & coarse & fine & paired drop & vs sibling & matching \\
\midrule
base model            & 0.047 & 0.031 & $+0.016$ \ci{-0.01}{+0.04} & 0.484 & 0.529 \\
G-inject (flow matching)   & 0.906 & 0.724 & $\bt{+0.182}$ \ci{+0.09}{+0.29} & 0.758 & 0.688 \\
U-inject (LM CE)           & 0.424 & 0.344 & $+0.081$ \ci{+0.02}{+0.16} & 0.706 & 0.925 \\
anchor @14 + flow matching & 0.966 & 0.888 & $+0.078$ \ci{+0.00}{+0.18} & \bt{0.914} & 0.854 \\
\bt{anchor @14 alone}      & 0.802 & \bt{0.747} & $+0.055$ \ci{-0.01}{+0.13} & \bt{0.878} & 0.721 \\
closed-form patch @ L7     & 0.323 & 0.307 & $+0.016$ \ci{-0.01}{+0.04} & 0.690 & 0.667 \\
\bottomrule
\end{tabular}
\end{center}
\end{table}

Evaluation coverage.
We scored the same generations with coarse banks, fine banks containing same-category
siblings, sibling-only comparisons, and CLIP's vision tower. The fine-grained evaluation uses 24
sibling pairs; the 56-concept sweep uses distinct-category banks. CLIP-based scores are $0.682$ for
G-inject and $0.844$ for anchoring plus flow matching.

\section{Protocol-level controls}\label{app:controls}

Table~\ref{tab:entrycontrols} collects the three controls \S\ref{sec:window} draws on. The
derangement row substitutes another concept's address at the same magnitude and direction
statistics, isolating the pairing from the perturbation; the two quarter-magnitude rows separate
dose from site.

\begin{table}[h]
\caption{Protocol-level controls for the entry-point sweep, group of eight, accuracy
($\uparrow$). The base model scores $0.254$ on matching and $0.036$ on identity.}
\label{tab:entrycontrols}
\begin{center}
\begin{tabular}{lccl}
\toprule
control & und.\ matching & gen.\ identity & evaluation role \\
\midrule
layer 3, real addresses & 0.887 & \bt{0.930} & reference condition \\
\bt{layer 3, deranged addresses} & 0.138 & \bt{0.000} & address-order control \\
layer 7, quarter magnitude & 0.613 & 0.086 & shallow-site magnitude control \\
layer 24, quarter magnitude & 0.138 & 0.031 & deep-site magnitude control \\
base model & 0.254 & 0.036 & unmodified reference \\
\bottomrule
\end{tabular}
\end{center}
\end{table}

Table~\ref{tab:protocolcontrols} collects five further controls on the same group: injection with
the name--image pairing shuffled, injection from text alone, embedding-only injection on either
side, a forgetting check on the underlying real categories, and open naming. No alternative route to
the binding comes near either trained condition, the real categories survive intact, and open naming
is the one column that separates the two injection directions. Cross-name retrieval, not tabulated,
gives $0.758$ for a name's own concept against $0.102$ for the other names of its group.

\begin{table}[h]
\caption{Protocol-level controls, group of eight, accuracy ($\uparrow$). Identity is 24-way
retrieval on generations and matching is the 4-way naming probe, both as in \S\ref{sec:metrics};
\emph{open naming} is the fraction of concepts for which the model volunteers the pseudo-word with
no candidates in the prompt. \emph{Real-category} is balanced accuracy on the underlying real
categories, the forgetting check. Cross-name retrieval is reported in the text.}
\label{tab:protocolcontrols}
\begin{center}
\begin{tabular}{lcccc}
\toprule
condition & identity & matching & open naming & real-category \\
\midrule
base model                          & 0.036 & 0.254 & 0.000 & 0.931 \\
\midrule
G-inject                            & 0.758 & 0.600 & 0.000 & 0.931 \\
U-inject                            & 0.562 & 1.000 & 1.000 & 0.924 \\
\midrule
name shuffling, G-inject            & 0.016 & 0.000 & 0.000 & --- \\
text-only injection                 & 0.055 & 0.238 & 0.000 & --- \\
embedding-only, generation side     & 0.172 & 0.238 & 0.000 & --- \\
embedding-only, understanding side  & 0.055 & 0.975 & 0.000 & --- \\
\bottomrule
\end{tabular}
\end{center}
\end{table}

\section{Reproducibility}\label{app:repro}

Every number in this paper is produced by a script in the accompanying code release, reading from a
results directory that the same scripts write. Concept rendering is deterministic given the asset
list and seed. Training runs are single-GPU and fully specified by their command lines, which are
generated by one experiment-matrix module rather than written by hand, so a stage can be re-run to
completion with one command and is idempotent per job. The aggregation scripts named in each section
regenerate the corresponding table from the raw per-condition JSON.

\end{document}